\documentclass[letterpaper,journal]{IEEEtran}
\usepackage{kennedy_style}
\usepackage{hyperref}
\usepackage{enumitem}
\setlist[itemize]{leftmargin=*}
\setlist[enumerate]{leftmargin=*}

\makeatletter
\def\IEEEbibitemsep{0pt plus 0.5pt}
\makeatother

\begin{document}
\title{J-PARSE: Jacobian-based Projection Algorithm for Resolving Singularities Effectively in Inverse Kinematic Control of Serial Manipulators}

\anonymoustext{\anonymous}{
\author{Authors omitted for review.
\thanks{Thanks omitted for review}}
}{ 
\author{Shivani Guptasarma$^1$, Matt Strong$^2$, Honghao Zhen$^1$, Monroe Kennedy III$^{1,2}$
    \thanks{This work was partly supported by Amazon Science. The second author was supported by NSF Graduate Research Fellowship DGE-2146755. The authors thank Oussama Khatib for course material containing the dynamic parameters of the PUMA560 manipulator and Alex Qiu for the software infrastructure used to train a diffusion policy through teleoperation of the X-Arm7. \\ 
    Project website: \url{https://jparse-manip.github.io/}. \\
    Authors are members of the Departments of $^1$Mechanical Engineering and $^2$Computer Science, Stanford University, Stanford CA, 94305. {\tt\small \{shivanig, mastro1, honghao\_zhen, monroek\}@stanford.edu}}  }
}


\maketitle 

\begin{abstract}
    J-PARSE is an algorithm for smooth first-order inverse kinematic control of a serial manipulator near kinematic singularities. The commanded end-effector velocity is interpreted component-wise, according to the available mobility in each dimension of the task space. First, a substitute ``Safety" Jacobian matrix is created, keeping the aspect ratio of the manipulability ellipsoid above a threshold value. The desired \new{velocity} is then projected onto non-singular and singular directions, and the latter projection scaled down by a factor informed by the \new{altered mobility}. A right-inverse of the non-singular Safety Jacobian is applied to the modified command. In the absence of joint limits and collisions, this ensures safe transition into and out of low-\new{mobility} configurations\new{, guaranteeing locally stable reaching behavior towards target poses within, on the boundary of, and outside the workspace. The behavior is further guaranteed to be locally asymptotically stable if the starting and target configurations are not exactly singular, even if they are nearly singular.} Velocity control with J-PARSE is benchmarked against approaches from the literature, \new{illustrating its use of a single tuning parameter to simultaneously achieve high reaching accuracy and stable behavior}. \new{A}pplications in \rev{teleoperation, }servoing, and learning \new{are demonstrated}.
\end{abstract}


\section{Introduction}

\IEEEPARstart{S}{everal} modern applications of robotic manipulators require the end-effector motion to be commanded in real time. Prominent among these are direct human operation (as in teleoperated or assistive robots controlled via joystick or other interfaces), \new{and} direct response to perception (as in visual servoing to a moving goal). \new{R}esponse to perception \new{may also  be} learned from human operation, as when \new{using task-space demonstrations given by a human operator to train}a planner, \new{to generate in real-time} an end-effector velocity conditioned on the perceived state of the environment. In all such applications, the approach taken to resolve \textbf{inverse kinematic \new{(IK)}} singularities impacts the accuracy and legibility of motion. The most common approach is that of simply avoiding singular regions. This not only reduces the \new{usable volume of the} workspace, but human operators -- even while they come to understand the limited mobility \new{near} singularities -- struggle to understand, and account for, the need to avoid such \new{configurations} altogether. \new{A}rticulated systems in nature, such as a human arm reaching out to its maximum extent, do not avoid singular configurations, even when their motion is based on a task-space goal.

This work proposes an algorithm, called\textbf{ J-PARSE (Jacobian-based Projection Algorithm for Resolving Singularities Effectively)}, to enable serial robotic manipulators to achieve externally-specified motion goals in the task space without \new{control discontinuities and reachability restrictions} arising from proximity to kinematic singularities. The algorithm is designed to slow down joint speeds, informed by the nature and extent of the decreased mobility in the vicinity of singularities. Its utility is demonstrated in providing stable task-space teleoperation and servoing throughout the workspace, as well as in ensuring that the end-effector velocities commanded by a perception-based real-time planning algorithm do not result in unstable computations or dangerously-high joint speeds. 

As human-in-the-loop control is variable, and unsuitable for extensive comparisons with other methods, a set of automated tasks is also presented\new{.} \new{The automated tasks include} either \emph{reaching} a discrete set of fixed target poses, or \emph{tracking} a slowly-moving target pose. These examples are used to illustrate the effects of varying the parameters involved in the algorithm, and the significance of each term in the proposed expression for joint speeds. Furthermore, \new{these examples show} that the manipulator is able to reach towards target poses outside the workspace, entering a singular configuration at the workspace boundary and later retreating from it to reach the next target pose. The tracking examples (both human-in-the-loop and automated) show that it is able to pass through the neighborhoods of internal singularities in a manner that allows the operator to tune the behavior, trading off deviation in direction \new{against maintenance of} speed. \new{This provides} an alternative to the default algorithms on commercial manipulators, where either such poses are avoided in the task space, or the extent of deviation is pre-programmed in proprietary code. 
\begin{figure}[!htbp] 
    \centering
    \includegraphics[width=\linewidth]{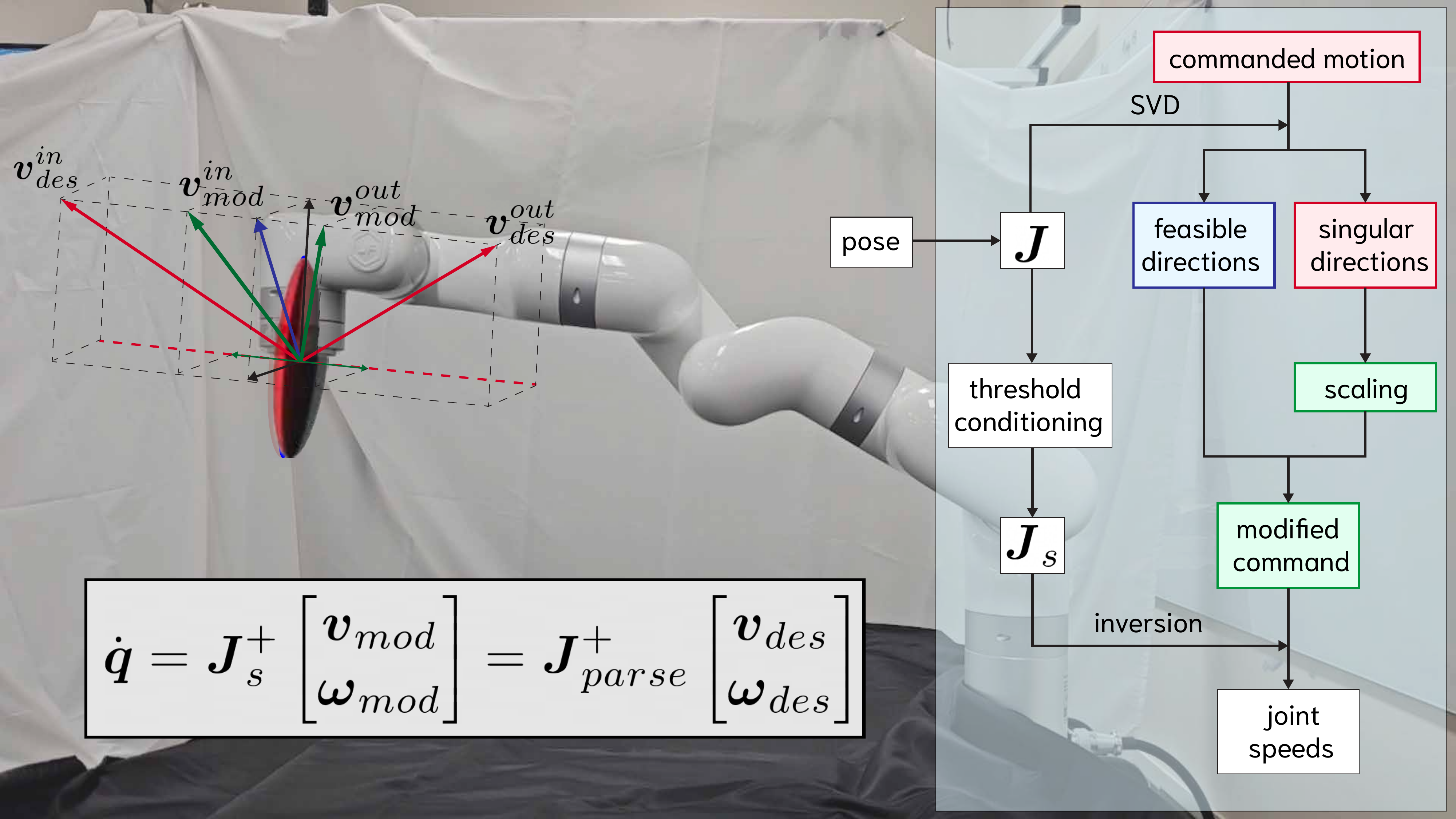} 
    \caption{
    By appropriately modifying the commanded velocity along singular directions, the manipulator is able to move both into and out of singular configurations. A safety Jacobian matrix~$\boldsymbol{J}_s$ is created by setting a minimum aspect ratio for the manipulability ellipsoid, preventing it from losing dimension and making it possible to invert~$\boldsymbol{J}_s$ to obtain a feasible set of joint speeds from the modified command.}
    \label{fig:jparse_splash} 
\end{figure}

An overview of the \del{method }\rev{approach} is shown in \figref{fig:jparse_splash}. The remainder of this paper is organized as follows. \secref{sc:lit} describes existing approaches relevant to addressing singularities in manipulator control. \secref{sc:prelims} explains mathematical preliminaries. \secref{sc:ps} \ins{describes the scope of the work, and the problem statement, framing the problem as one of reaching the \emph{next desired} end-effector pose, given the input of a task-space twist aiming to reach that pose}. Details of the proposed algorithm and its implementation are provided in \secref{sc:mainalg}. \secref{sc:contproof} proves \new{locally }stable reaching \textit{of} poses \new{in the interior and at the boundaries of} the workspace, and \textit{towards} poses outside the workspace, through a Lyapunov analysis. Stability-related considerations for discrete time implementation are provided in \secref{sc:discrete}, \new{based on a first-order approximation. S}ome variations \new{of the algorithm are} proposed in \secref{sc:mods}. \secref{sc:results} discusses the outcomes of implementing J-PARSE on various simulated and physical systems, with a brief discussion in \secref{sc:disc} placing the behavior in context. The approach is summarized and avenues for future work are mentioned in \secref{sc:conc}. 

\section{Related Work}
\label{sc:lit}
At a kinematic singularity, a serial manipulator loses one or more \gls{DoF}. The geometric Jacobian matrix, which maps actuator speeds to the linear and angular velocities of the end-effector, becomes increasingly ill-conditioned in the vicinity of a singular configuration, until it loses rank at the singularity. Singularities are subsets of the configuration space, and may occur at the boundary of the physical workspace or in its interior.  

In structured environments, it is often feasible to avoid singular configurations through offline considerations in goal specification and path planning. It is also desirable, in industrial settings, to avoid singularities so that the end-effector is able to traverse the planned path at a high speed throughout. However, singularity-avoidant path planning or workspace design is not always feasible during human-in-the-loop control, or online tracking of external moving targets; \ins{nor are high-speed and industrial precision typically priorities}. In these scenarios, it is \ins{merely preferable}, at every instant, to move the end-effector in a manner \new{that is }as close as possible to the commanded direction. 

\new{S}tandard teleoperation algorithm\new{s} in commercially-available manipulator arms~\cite{kinova_gen2_manual, xarm_manual}, and open-source servoing packages such as MoveIt Servo and MoveIt~2~Servo~\cite{moveit, moveit_servo, moveit_2_servo}), \new{rely upon \textit{singularity avoidance}. This }entails simply stopping (or slowing down and stopping) the motion near a singularity. Virtual potential fields have been used to repel the teleoperated motion away from singularities~\cite{khatibRealtimeObstacleAvoidance1985}. Some approaches also give feedback to the user in such situations~\cite{maneewarn1999augmented}. \new{S}uch an approach limits the \new{usable workspace} of the manipulator \new{that is already constrained by considerations of joint limits and collision avoidance. Naturally, it} can cause frustration to human operators~\cite{edls2017}. 

\new{Rather than designating singular neighborhoods as unusable, another approach} is to switch to joint space control near singularities. As long as an \new{IK} solution exists at the target pose, motions of individual joints can be executed via interpolation~\cite{xarm_manual}. As expected, when transmitted through the non-linear forward kinematic map, these may result in unintuitive motions \new{of} the end-effector. \ins{Moreover, numerical iterative methods of performing static (zeroth-order) \new{IK} also suffer from the effects of singularities (especially when the target pose itself corresponds to a singular configuration). \new{They therefore} need to incorporate the same damping approaches (described further below) that exist for singularity handling in first-order \new{IK} control~\cite{quik2022}.}

In operational space force control, control at and near singularities has been demonstrated by treating a singular configuration as a redundant one~\cite{changManipulatorControlKinematic1995}. The principal axes of the velocity ellipsoid corresponding to the singular (lost) directions can be dropped to obtain a redundant Jacobian matrix (referred to in the present work as the \emph{projection} Jacobian matrix). By appropriately selecting a vector in the nullspace of this matrix, informed by the dynamic model, and designing a potential function in the joint space, the robot can be moved into and out of singular configurations. This approach relies on analysis of the types of singularity, and appropriate selection of the potential functions, but also accurate knowledge of the dynamic model of the manipulator and payload, which is not always feasible~\cite{peters2006}. 

First-order \new{IK} control too has been studied in the operational space, with an emphasis on eliminating the discontinuity that arises in least-squares inversion (using the pseudoinverse) of the Jacobian matrix during task activation and deactivation, by recursively defining a continuous inverse~\cite{mansard2009a}. This continuous inverse is sensitive to the singularities of the Jacobian matrix, and still requires additional smoothing using Damped Least-Squares~\cite{mansard2009b}. Alternatively, it has been proposed that singularity avoidance be framed as a unilateral constraint and integrated into the framework for velocity or torque control~\cite{mansard2009b}. The observation that the pseudoinverse causes a discontinuity is very pertinent to the discussion of \new{IK} control near singularities. However, \new{smoothing} control by treating motion along the singular direction as a ``deactivated" task prevent\new{s} the manipulator from reaching a target pose close to singularity~\cite{han2012, han2013}. 

The most well-known approach for smooth \new{IK} control near singularities is the \gls{DLS} inversion of the Jacobian matrix, which adds a small positive value to the otherwise vanishing denominators during inversion~\cite{nakamuraInverseKinematicSolutions1986, wamplerManipulatorInverseKinematic1986}. \gls{DLS} avoids numerical instability when the Jacobian becomes ill-conditioned, but does so at the cost of accuracy, even affecting the accuracy at non-singular poses where it is well-conditioned~\cite{deo1995}. The numerical modifications comprising the \gls{DLS} approach result in motion that is not only inexact, but may also be unintuitive for human operators~\cite{edls2017, carmichael2020}. Several variations of \gls{DLS} have been proposed to mitigate \del{this} \rev{its} inexact behavior~\cite{deo1995, edls2017}. \new{Together, these may be referred to as \textbf{Adaptively-Damped Least Squares or ADLS}.} For example, the damping factor may be adjusted based on proximity to singularity~\cite{nakamuraInverseKinematicSolutions1986}, the rate of approach to or departure from singularity~\cite{kelmar1988}, or the distance to the target~\cite{chan1988}. The damping may also be applied selectively to different singular values of the Jacobian matrix, based on the difficulty in reaching the target~\cite{Buss2005SelectivelyDL}. These adaptive implementations require careful empirical tuning of multiple parameters. In a relatively recent critical review of \gls{DLS} algorithms with various implementations of adaptive damping selection~\cite{devito2017}, the authors of the study concluded that \textbf{\textit{none} of the \new{ADLS }algorithms compared were able to provide satisfactory robot-independent and trajectory-independent stable behavior}. For example, the goal pose being outside the reachable workspace would result in chattering, or the choice of non-zero damping too far away from singularities would result in \new{steady-state} errors. \new{Furthermore}, none of these \rev{approaches }for adaptive gain selection addressed the challenge of interpretability and intuitiveness. 

In order not to disturb exactness of control away from singularities, while slowing motion near singularities \rev{\emph{in a manner that appears intuitive to a human operator}}, a tunable \textbf{\gls{EDLS}} framework has been proposed~\cite{edls2017}. Damping is framed as a function of the singular values, so that motion is damped particularly in those directions where mobility is lost. \new{N}on-singular directions remain practically undamped, while\new{,} when the pose is nearly singular, the singular direction is completely damped (velocity brought to zero). The damping is asymmetrically applied, impeding motion along the singular direction while approaching a singularity, but encouraging it while departing. Artificial potential fields are also applied to repel the teleoperator from singular poses. \new{Whereas singularity handling algorithms had traditionally prioritized }trajectory tracking performance, \new{t}he \gls{EDLS} framework represents an important shift of focus towards the user experience of a human in the loop. \gls{EDLS} aims to make teleoperation intuitive near singularities and away from them~\cite{carmichael2020}\new{. However,} \rev{it} \textbf{disallows entering the immediate neighborhood of a singular configuration}. That is, EDLS does not aim to safely and smoothly operate within, and across the boundary of, a singular neighborhood -- even though a human user, while acknowledging the necessary constraints on speed and accuracy, may still wish to do so.

In summary, \rev{first-order} \new{IK} based methods are widely used to control the end-effector motion of serial manipulators, as they do not require global analysis of kinematics \rev{(\textit{a-priori} identification of all singular configurations)}, accurate identification of dynamics, prior specification of targets, or deliberate adjustment of the workspace. \new{However, such methods suffer from the problem of IK} singularities. \new{Control near singularities} relies on either solving the zeroth-order \new{IK} \ins{in closed form} and switching to joint space, explicitly enforcing singularity avoidance as a task objective, or numerical modifications along the lines of \new{A}DLS. \new{These existing approaches }prevent accurate reaching of target poses in general, rely on complicated parameter selection approaches, and do not account for user \new{or designer} experience. Even in the most recent approach prioritizing user experience in the proximity of singularities~\cite{carmichael2020}, the immediate neighborhood of the singularity is actively avoided. Among popular commercial arms, the UFactory X-Arm~\cite{xarm_manual} hits its emergency stop as it approaches singularity. The Kinova Gen3 controller allows the end-effector to \ins{deviate from the commanded direction during teleoperation near both internal and boundary singularities~\cite{kinova_gen3_manual}, but the extent of deviation cannot be specified or tuned}. To the best of our knowledge, there does not exist an \new{easily and reliably tunable} algorithm for \new{IK} control prioritizing stable control at, near and away from singularities. The goal of this work is to propose such an algorithm, while deferring treatment of joint limits and collision constraints to future work. 

\new{
\begin{table}[h]
    \centering
    \caption{The pose is represented by~$\boldsymbol{p}$, the twist (linear and angular velocities) by~$\boldsymbol{\mathcal{V}}$, and damping by~$\lambda$. The inherent loss of mobility near singularities prevents any algorithm from accurately matching arbitrary velocities in the task space. Poorly-tuned ADLS shows the same limitations as DLS. J-PARSE achieves safety and accurate reaching of poses throughout the reachable workspace, with only one parameter to be tuned.}
    \label{tab:qualcomp}
    \begin{tabular}{l|c|c|c|c|c|c|c}
        \textbf{Algorithm} & \rotatebox{90}{\makecell[l]{Accurate~$\boldsymbol{p}$ \\ Away from sings.}} & \rotatebox{90}{\makecell[l]{Accurate~$\boldsymbol{p}$ \\ Near sings.}} & \rotatebox{90}{\makecell[l]{Accurate~$\boldsymbol{\mathcal{V}}$ \\ Away from sings.}} & \rotatebox{90}{\makecell[l]{Accurate~$\boldsymbol{\mathcal{V}}$ \\ Near sings.}} & \rotatebox{90}{\makecell[l]{Stability \\ Near sings.}} & \rotatebox{90}{\makecell[l]{Overall safety}} & \rotatebox{90}{No. of params.} \\
        \hline
        (Pseudo/)inv. & \cellcolor[HTML]{ccebc5}\cellcolor[HTML]{ccebc5}$\checkmark$ & $\times$ & \cellcolor[HTML]{ccebc5}$\checkmark$ & $\times$ & $\times$ & $\times$ & \cellcolor[HTML]{ccebc5}$0$ \\
        Avoidance & \cellcolor[HTML]{ccebc5}$\checkmark$ & $\times$ & \cellcolor[HTML]{ccebc5}$\checkmark$ & $\times$ & -- & \cellcolor[HTML]{ccebc5}$\checkmark$ & \cellcolor[HTML]{ccebc5}$0$ \\
        DLS (high~$\lambda$) & $\times$ & $\times$ & $\times$ & $\times$ & \cellcolor[HTML]{ccebc5}$\checkmark$ & \cellcolor[HTML]{ccebc5}$\checkmark$ & \cellcolor[HTML]{e6f5e2}$1$ \\
        DLS (low~$\lambda$) & \cellcolor[HTML]{ccebc5}$\checkmark$ & $\times$ & \cellcolor[HTML]{ccebc5}$\checkmark$ & $\times$ & $\times$ & $\times$ & \cellcolor[HTML]{e6f5e2}$1$  \\
        ADLS (tuned) & \cellcolor[HTML]{ccebc5}$\checkmark$ & \cellcolor[HTML]{ccebc5}$\checkmark$ & \cellcolor[HTML]{ccebc5}$\checkmark$ & $\times$ & \cellcolor[HTML]{ccebc5}$\checkmark$ & \cellcolor[HTML]{ccebc5}$\checkmark$ & \cellcolor[HTML]{f0faf0}$2$ \\
        EDLS & \cellcolor[HTML]{ccebc5}$\checkmark$ & $\times$ & \cellcolor[HTML]{ccebc5}$\checkmark$ & $\times$ & -- & \cellcolor[HTML]{ccebc5}$\checkmark$ &$3$ \\
        J-PARSE & \cellcolor[HTML]{ccebc5}$\checkmark$ & \cellcolor[HTML]{ccebc5}$\checkmark$ & \cellcolor[HTML]{ccebc5}$\checkmark$ & $\times$ & \cellcolor[HTML]{ccebc5}$\checkmark$ & \cellcolor[HTML]{ccebc5}$\checkmark$ & \cellcolor[HTML]{e6f5e2}$1$  \\
        \hline
    \end{tabular}
\end{table}
The proposed algorithm, J-PARSE, provides a \textbf{mathematical guarantee of locally stable motion} throughout the workspace for any serial manipulator, including \textbf{inside and through singular neighborhoods}, without requiring any dynamic model or imposing artificial restrictions to the usable workspace, and with \textbf{a single, kinematically-meaningful, parameter to be tuned}. The algorithm uses no additional inputs in real-time except those already used in IK control, and is agnostic to the architecture of the manipulator. Table~\ref{tab:qualcomp} summarizes its features in comparison to other approaches.}

The potential applications of the proposed approach lie not only in teleoperation and servoing, but \new{also in learning of autonomous policies}.  Learning policies \new{with J-PARSE enables} robots to interact with objects at poses that can only be reached by assuming near-singular configurations. State-of-the-art \new{robot learning approaches predict actions as task-space velocities, but without addressing singularities~\cite{black2026pi0visionlanguageactionflowmodel, kim2024openvla, paloKeypointActionTokens2024}, or switch to position rather than velocity control in the task space~\cite{chiDiffusionPolicyVisuomotor2024}, or move to the joint space to avoid the issue~\cite{zhaoLearningFineGrainedBimanual2023}}. Works in robot learning that do address singularit\new{ies} avoid \new{them} explicitly by encouraging the robot to avoid areas in the task space with a low manipulability measure~\cite{bhardwaj2022storm, manavalan2019learning, ding2024bunny}. J-PARSE makes it safe to collect data for \new{training} robotic policies in the \new{vicinity of singular configurations}, and trained policies using it are able to autonomously complete tasks that demand near-singular configurations.

\section{Preliminaries}
\label{sc:prelims}
This section gives an overview of first-order \new{IK} control, and the problem of singularities. 
\subsection{Forward kinematics}
\label{sc:fk}
Consider a serial robotic manipulator with~$n$ \rev{actuators, whose actuator positions are denoted by the vector~$\boldsymbol{q}_{n \times 1}$. The manipulator's end-effector pose may be denoted as~${\boldsymbol{p} = \begin{bmatrix}x & y & z & \theta_x & \theta_y & \theta_z \end{bmatrix}^\top} \in \new{\reals{6}}$,~\ins{using a local parameterization of~$\sethree$}, where we define~${\hat{\boldsymbol{h}} \theta = \begin{bmatrix}\theta_x & \theta_y & \theta_z \end{bmatrix}^\top}$ using the axis angle representation. That is,~$\hat{\boldsymbol{h}}$ is a unit vector pointing along the axis of rotation, with respect to a fixed reference frame, and~${\theta \in [0, \pi]}$ is the counter-clockwise angle of rotation about~$\hat{\boldsymbol{h}}$, such that \rev{the rotation matrix transforming the reference frame into the end-effector frame is given by\footnote{\rev{where ~$\hatmap{(\cdot)}$  denotes the skew-symmetric matrix operator, and~$\exp(\cdot)$ denotes the exponential map from the Lie algebra~$so(3)$ to the special orthogonal group~$SO(3)$.}}~$\boldsymbol{R} = \exp(\hatmap{\hat{\boldsymbol{h}} \theta})$}}.

The linear velocity~$\boldsymbol{v}$ and angular velocity~$\boldsymbol{\omega}$ of the end-effector relate to the actuator speeds via the geometric Jacobian matrices: 
\begin{align}
&\boldsymbol{v} = \begin{bmatrix} \left(\pd{x}{\boldsymbol{q}}\right)^\top & \left(\pd{y}{\boldsymbol{q}}\right)^\top & \left(\pd{z}{\boldsymbol{q}}\right)^\top \end{bmatrix}^\top \dot{\boldsymbol{q}}
= \rev{\boldsymbol{J}_v \dot{\boldsymbol{q}}} \\
&\hatmap{\boldsymbol{\omega}} = \dot{\boldsymbol{R}} \boldsymbol{R}^\top = \rev{\left(\sum_{i=1}^{\new{n}} \pd{\boldsymbol{R}}{q_i} \dot{q_i} \right) \boldsymbol{R}^\top}, \label{eq:omegacalc} \\
&\rev{\boldsymbol{\omega}} = \rev{\boldsymbol{J}_\omega \dot{\boldsymbol{q}}}  
\end{align}
which may be written together as \rev{the twist:}
\begin{equation}
    \boldsymbol{\mathcal{V}} = \begin{bmatrix} \boldsymbol{v} \\ \boldsymbol{\omega} \end{bmatrix} = \begin{bmatrix}
        \boldsymbol{J}_v \\ \boldsymbol{J}_\omega
    \end{bmatrix} \dot{\boldsymbol{q}}. 
\end{equation}
\new{For dimensional consistency, the columns of~$\boldsymbol{J}_v$ corresponding to revolute joints have dimensions of length and those corresponding to prismatic joints are dimensionless. The columns of~$\boldsymbol{J}_\omega$ corresponding to revolute joints are dimensionless and those corresponding to prismatic joints have inverse length dimensions.} \rev{In general, for an~$m$-dimensional task space, the first-order forward kinematic relationship is\footnote{\rev{Throughout this article, in order to avoid confusion with other subscripts, vectors are treated as one-dimensional matrices for representing their dimension, e.g.,~$\boldsymbol{\mathcal{V}}_{m\times 1} \in \reals{m}$. Matrices are represented in upper case and vectors\new{, except~$\boldsymbol{\mathcal{V}}$,} in lower case. The~$(i, j)^{th}$ element of matrix~$\boldsymbol{A}$ is denoted as~$A_{ij}$ in general and~$A_i$ is a shorthand for~$A_{ii}$.}}:}
\begin{equation}
    \rev{\boldsymbol{\mathcal{V}}_{m\times 1} = \boldsymbol{J}_{m \times n}(\boldsymbol{q})~\dot{\boldsymbol{q}}_{n \times 1}} \label{eqn:jacobian_definition}
\end{equation}
The \rev{mobility }\del{ability of the manipulator to translate along and rotate about directions in the task space} is numerically characterized by the \gls{SVD} of~$\boldsymbol{J}$:
\begin{equation}
    \boldsymbol{J}_{m \times n} = \boldsymbol{U}_{m \times m} \boldsymbol{\Sigma}_{m \times n} \boldsymbol{V}^\top_{n \times n}
    \label{eqn:J_svd}
\end{equation}
where the columns of $\boldsymbol{U}$ are the left singular-vectors of $\boldsymbol{J}$ and correspond to the principal directions in which the end-effector can move, $\boldsymbol{\Sigma}$ contains the singular values, and the columns of $\boldsymbol{V}$ are the right singular-vectors relating to motion in the joint space. \del{For non-redundant manipulators the task and joint space dimension have the same dimension: ${m = n}$.} For a redundant manipulator,~${m < n}$; \rev{otherwise,~${m = n}$.} In general,~${\boldsymbol{\Sigma}_{m \times n}}$ is a block diagonal matrix: 
\begin{equation}
    \boldsymbol{\Sigma}_{m \times n} = \begin{bmatrix}
        \begin{bmatrix}
            \sigma_1 &  \ldots & 0 \\
            0 & \ddots & 0 \\
            0 & \ldots & \sigma_m
        \end{bmatrix}_{m \times m}, & \boldsymbol{0}_{m \times (n-m)}
    \end{bmatrix},
    \label{eqn:expanded_singular_matrix}
\end{equation}
where\footnote{\new{Singular values are non-negative by definition. The non-increasing ordering is a convention in SVD.}}~$\new{\sigma_1 \geq \sigma_2 \geq \dots \sigma_m} \geq 0$, for~$ i = 1, \dots, m$.

The principal axes of the manipulability ellipsoid (which contains the set of all possible end-effector velocities for a unit norm~$\dot{\boldsymbol{q}}$) can be constructed by using the orthogonal singular vectors in~$\boldsymbol{U}$ scaled by the corresponding singular values~$\sigma_i$. 

\new{To ensure consistent physical dimensions, the elements of~$\boldsymbol{U}$ correspond to linear or angular task space DoFs (and account for the units of~$\boldsymbol{v},~\boldsymbol{\omega}$), elements of~$\boldsymbol{V}^\top$ correspond to revolute or prismatic joints (accounting for the units of~$\dot{\boldsymbol{q}}$), and the \textbf{singular values have no physical units}.}

\subsection{Zeroth-order Inverse kinematics}
\new{
The zeroth-order IK problem is that of finding~${\boldsymbol{q} \in \mathbb{C}^n}$ (and the real subset\footnote{\new{IK solutions are complex in general. None, some, or all of them might be real, depending upon~$\boldsymbol{p}$ and the architecture.}}~${\boldsymbol{q} \in \mathbb{R}^n}$), such that the end-effector pose achieves a given pose~${\boldsymbol{p} \in \reals{m}}$. In a redundant manipulator, since~${m < n}$, infinitely many IK solutions exist. Various optimality conditions may be introduced in order to resolve the redundancy~\cite{nakamura1990advanced}. However, even when~${m = n}$, the IK solution, and therefore the robot posture, is not unique. For example, a planar~$2$-R manipulator has two IK solutions, the so-called elbow-up and elbow-down configurations~\cite{schreiber2022determination}, both leading to the same end-effector position~${\boldsymbol{p} \in \reals{2}}$. A general $6$-R serial manipulator in space has~$16$ IK solutions~\cite{raghavan1993inverse}, for a given~${\boldsymbol{p} \in \reals{6}}$ representing a pose in~$\sethree$. It may or may not be possible to obtain all the solutions in analytical form for a given manipulator. However, the existence of multiple IK solutions is a result of the generally non-injective nature of the forward kinematic map\footnote{\new{unless all joints are prismatic}}, and is true of both non-redundant and redundant manipulators. 

}

\subsection{First-order inverse kinematics}
\label{sc:1ik}

A manipulation task may require the end-effector to have a specified\del{ linear and angular velocity (twist) in the task space, written as}~$\boldsymbol{\mathcal{V}}_{\mathrm{des}}$. It then becomes necessary to find~$\dot{\boldsymbol{q}}$ satisfying~\eqref{eqn:jacobian_definition}. \new{If~$\boldsymbol{J}$ is square (the manipulator is non-redundant) and has full rank, then it is exactly invertible and the solution is unique. }In a redundant manipulator,~$\boldsymbol{J}$ is a rectangular matrix \rev{with no proper inverse}. \new{Therefore, i}f it \new{has full rank}, there exist infinitely many solutions for~$\dot{\boldsymbol{q}}$. Any right-inverse of~$\boldsymbol{J}$ gives a feasible solution. The Moore-Penrose inverse, commonly known as the regular pseudoinverse, yields the solution with the least-squared norm:
\begin{align}
    \dot{\boldsymbol{q}}_{\mathrm{des}, n \times 1} &= \boldsymbol{J}^{+}_{\rev{n \times m}} \boldsymbol{\mathcal{V}}_{\mathrm{des, }m \times 1} = \new{\argmin_{\dot{\boldsymbol{q}},~\mathrm{subj.to}~\boldsymbol{J}\dot{\boldsymbol{q}} = \boldsymbol{\mathcal{V}}_{\mathrm{des}}}||\dot{\boldsymbol{q}}||^2}.
    \label{eqn:basic_Jpinv_definition}
\end{align}
In this work, unless specified otherwise,~$\boldsymbol{J}^{+}$ denotes the right Moore-Penrose inverse \rev{or pseudoinverse}. However, other generalized inverses exist, which optimize other objectives. 

Numerically, the pseudoinverse is calculated using the \gls{SVD}. \new{If~$\boldsymbol{J}$ has full rank}:
\begin{equation}
    \boldsymbol{J}^{+} = \boldsymbol{J}^\top (\boldsymbol{J} \boldsymbol{J}^\top)^{-1} = \boldsymbol{V} \boldsymbol{\Sigma}^{+} \boldsymbol{U}^\top,
    \label{eqn:Jpinv_svd}
\end{equation}
where 
\begin{equation}
    \Sigma^{+}_{n \times m} = \begin{bmatrix}
        \begin{bmatrix}
            \frac{1}{\sigma_1} &  \ldots & 0 \\
            0 & \ddots & 0 \\
            0 & \ldots & \frac{1}{\sigma_m}
        \end{bmatrix}_{m \times m} \\ \boldsymbol{0}_{(n-m) \times m}
    \end{bmatrix},
    \label{eqn:expanded_singular_matrix_inverse}
\end{equation}
as, for \new{full-rank}~$\boldsymbol{J}$, all the singular values are non-zero. 

\subsection{Singularities}
\label{sc:singps}
\new{The matrix}~$\boldsymbol{J}$ is singular iff one or more of the singular values vanish:~${\prod_{i=1}^m\sigma_i = 0}$. Equivalently, the rank of~$\boldsymbol{J}$ falls below~$m$. Then \eqref{eqn:jacobian_definition} has infinitely many solutions when~${\boldsymbol{\mathcal{V}} \in \mathcal{C} (\boldsymbol{J})}$ (the column space of~$\boldsymbol{J}$), but no solution otherwise. The pseudoinverse is also defined for a singular matrix. In this case~$\boldsymbol{J}^+$ results in the least-squared error solution:
\begin{align}
    \dot{\boldsymbol{q}}_{\mathrm{des}, n \times 1} = &\boldsymbol{J}^{+}_{n \times m} \boldsymbol{\mathcal{V}}_{\mathrm{des, }m \times 1} \\ = &\new{\argmin_{\dot{\boldsymbol{q}} \in \mathcal{M}}||\dot{\boldsymbol{q}}||^2} \\
    &\new{\mathrm{where} \quad \mathcal{M} = \argmin_{\boldsymbol{z}}||\boldsymbol{\mathcal{V}}_{\mathrm{des}} - \boldsymbol{J} \boldsymbol{z}||^2}
    \label{eqn:sing_Jpinv_definition}
\end{align}
For a singular matrix, the matrix~$\boldsymbol{J} \boldsymbol{J}^\top$ in \eqref{eqn:Jpinv_svd} is not invertible. Instead, the pseudoinverse is found by replacing~$\frac{1}{\sigma_i}$ by~$0$ in \eqref{eqn:expanded_singular_matrix_inverse}, wherever~$\sigma_i = 0$. Clearly, the above definition is discontinuous with the definition of pseudoinverse for non-singular matrices:
\begin{equation}
    \lim_{\sigma_i \rightarrow 0^+}\left(\frac{1}{\sigma_i}\right) = \infty \ne 0, \nonumber
\end{equation}
\rev{and, therefore, is not practically useful in manipulator control, as }
\textbf{in practice,~$\boldsymbol{J}$ is not singular or non-singular in a discrete sense; rather, it becomes progressively ill-conditioned as the manipulator approaches a singular configuration}. Approaching a singularity corresponds to $\sigma_i \rightarrow 0^+$. 

\new{\textbf{Measures of proximity}: }The distance from singularity is captured through a variable $\kappa$ as either the manipulability measure:
\begin{equation}
    \kappa_{\mathrm{mm}} = \sqrt{\det(\boldsymbol{J} \boldsymbol{J}^\top)} = \prod_{i=1}^m\sigma_i
    \nonumber
\end{equation}
which is $\kappa_{\mathrm{mm}} = 0$ at a singularity, or through the condition number \new{(ratio of largest to smallest singular value)}:
\begin{equation}
     \kappa_{\mathrm{cn}} = \frac{\sigma_{\mathrm{max}}}{\sigma_{\mathrm{min}}} \geq 1, 
     \label{eqn:condition_number}
\end{equation}
for which~$\kappa_{\mathrm{cn}} \rightarrow \infty$ at a singularity. \new{Here, we use the inverse condition number (ICN)~$\frac{1}{\kappa_{\mathrm{cn}}}$, which equals~$0$ at a singularity}. The larger the condition number and smaller the manipulability measure, the more ill-conditioned~$\boldsymbol{J}$ \new{becomes}.

\new{\textbf{Singularities and IK solutions}: By the contrapositive of the Inverse Function Theorem or IFT~\cite{rudin1976principles}, if two or more IK branches (continuous families of solutions) meet at a configuration, then the forward kinematic map intersects itself there and~$\boldsymbol{J}$ is singular. When exiting such a configuration, it is important to select an appropriate IK branch\footnote{\new{In some manipulators, called \textit{cuspidal} manipulators, the robot may arrive at a different IK solution without having passed through a singular configuration~\cite{wengerReviewCuspidalSerial2022}. Well-defined unique branches do not exist in such manipulators, as the same continuous family in the joint space may traverse a task-space pose more than once at two different postures (joint configurations)~\cite{salunkheKinematicIssues6R2025}. Postures are, however, split into \textit{aspects}, which, as in non-cuspidal manipulators, meet at singularities~\cite{wengerReviewCuspidalSerial2022}.}} from the available alternatives~\cite{schreiber2022determination}.
}

\new{\textbf{Issues in inversion}: }When approaching the singularity, using the Jacobian pseudoinverse as in \eqref{eqn:expanded_singular_matrix_inverse} causes numerical instability and an erratic motion, as a denominator in $\boldsymbol{J}^{+}$ approaches zero and results in high commanded joint speeds. \delwhole{Instead, in order to limit the commanded joint speeds, a small region may be defined in the neighborhood of the singularity, within which the matrix is treated as singular by switching to the definition in \eqref{eqn:sing_Jpinv_definition}. In this case, division by very small numbers is avoided, but there is a discontinuity in commanded joint speeds when the control\rev{ler} \del{method }is changed, resulting in very high acceleration. }
\rev{The most common resolution is to solve a \emph{damped} least-squares optimization in the vicinity of singularities, instead computing an inexact solution to \eqref{eqn:jacobian_definition}:
\begin{equation}
    \dot{\boldsymbol{q}}_{\mathrm{des}, n \times 1} =  \mathrm{argmin}_{\dot{\boldsymbol{q}}}||\boldsymbol{\mathcal{V}}_{\mathrm{des}} - \boldsymbol{J} \dot{\boldsymbol{q}}||^2 + \lambda^2 ||\dot{\boldsymbol{q}}||^2,
\end{equation}
where~$\lambda$ may be varied as described in \secref{sc:lit}; it causes inaccuracy if too high, and instability if too low.} Singularity is thus a fundamental problem in \rev{first-order} \new{IK} control. 


\section{Problem Statement}
\label{sc:ps}
\subsection{Scope}
\label{sc:scope}
The goal of this work is to develop an algorithm that allows a manipulator under first-order \new{IK} control to \new{(a)} reach target poses corresponding to near-singular and singular configurations, \new{(b) extend as far as possible} towards unreachable targets, and, \new{(c)} retract from a near-singular configuration\new{, all in a stable manner}. 

The incorporation of considerations specific to the robot and application is left for future work. These include (a)~identifying a feasible path from the current to the desired pose, free of collisions with the environment, joint position limits, link collisions, or intervening workspace boundaries\footnote{Paths that appear to be internal to the reachable workspace may in fact pass through or culminate in unreachable poses. In cuspidal robots (e.g., the spatial~$3$-R manipulator studied in~\cite{wengerReviewCuspidalSerial2022}), as the locations of singular surfaces encountered in the task space depend upon the \new{IK posture}, there may be otherwise-\new{feasible paths} that are simply \new{infeasible} from the current (starting) joint space configuration. In non-cuspidal robots, the workspace may still be non-convex and contain voids \new{or self-intersections of its boundaries}. The challenge of \emph{reaching} such ``internal" boundaries is distinct from that of reaching poses that lie beyond them. The latter is an aspect of the broader problem of checking the existence of a feasible path to the goal, and is not the focus of the current work.}, and (b)~selection of the most appropriate \new{IK} branch while exiting a singular \new{configuration}, based on the state of the environment, the known properties of the manipulator, and the motion desired from it in the near future. 

In this work, the problem of tracking a moving target is treated as a sequential reaching task. Furthermore, the problem of following an externally-specified end-effector velocity is treated as equivalent to placing the end-effector under proportional control for reaching towards the desired ``next'' pose.

As joint speeds cannot be infinitely large, it is physically impossible to achieve an arbitrary end-effector velocity \emph{exactly} in a near-singular neighborhood. Any singularity-handling approach would involve either scaling down the desired speed, deviating from the desired direction, or both. In a real-time setting with a human in-the-loop, such choices are well-motivated; that is, slowly but legibly approaching singular configurations is preferable to stopping at a pre-set distance away from them. Unlike an offline motion planner, a human operator continuously adapts their input to the observed behavior of the manipulator. Hence, the emphasis in this work is placed on reachability and tunability of behavior, rather than on time taken, following the example set by previous work in this area~\cite{edls2017}.

\subsection{Formulation}
\label{sc:forml}
\ins{Assuming a desired end-effector twist~$\boldsymbol{\mathcal{V}}_{\mathrm{des}}$ provided in real time by an external source, such as a teleoperating human in the loop, the goal of this work is to ensure that the manipulator can be \new{locally }stably controlled, everywhere in its reachable workspace. Equivalently, the goal is to prove that the next desired end-effector pose~$\new{\boldsymbol{p}_{\mathrm{des}}}$}, offset slightly from the current pose in the direction of~$\boldsymbol{\mathcal{V}}_{\mathrm{des}}$, is stably reachable. 

For a desired target pose $\boldsymbol{p}_{\mathrm{des}}$ \new{in the interior, exterior or boundary of the workspace}, a successful control\rev{ler} \del{method }minimizes the error between~$\boldsymbol{p}_{\mathrm{des}}$ and the final pose~$\lim_{t \rightarrow \infty} \boldsymbol{p}(t)$. As~${\boldsymbol{p}_{\mathrm{des}}, \boldsymbol{p}(t) \in \new{\reals{6}}}$, error norms are defined separately for position and orientation. An error vector is defined as:
\begin{align}
    \boldsymbol{e}: \new{\reals{6}} \times \new{\reals{6}} \nonumber \rightarrow \new{\reals{6}} \\
    \boldsymbol{e}(\boldsymbol{p}_1, \boldsymbol{p}_2) = \begin{bmatrix} k_{\mathrm{pos}} (\boldsymbol{x}_1 - \boldsymbol{x}_2) \\ k_{\mathrm{ori}} \theta_{12} \boldsymbol{h}_{12}\end{bmatrix}, \label{eq:errordef}
\end{align}
where
\begin{itemize}   
    \item $k_{\mathrm{pos}}, k_{\mathrm{ori}} \rev{> 0}$ are weights for position and orientation errors \rev{(which can be selected based on priority and judgment, as the physical dimensions of linear and angular errors are mismatched in any case),}
    \item ${\boldsymbol{x}_{\rev{i}} = \begin{bmatrix}x_{\rev{i}} & y_{\rev{i}} & z_{\rev{i}} \end{bmatrix}^\top}$, \rev{where~$i \in \{1, 2\}$}, and
    \item $\theta_{12} \hat{\boldsymbol{h}}_{12}$ is the axis-angle representation of the relative rotation~$\boldsymbol{R}_1 \boldsymbol{R}_2^\top$ between the two orientations, with the rotation matrices~$\boldsymbol{R}_1, \boldsymbol{R}_2$ corresponding to the orientations of~$\boldsymbol{p}_1, \boldsymbol{p}_2$.
\end{itemize}
The vector~$\boldsymbol{e}$ is inhomogeneous (combining components \new{representing displacements in both, the Euclidean space of positions~$\reals{3}$ and the non-Euclidean space of orientations~$\sothree$}). \new{Still, in practice,} the task of reaching a goal pose can be framed as a minimization of \del{its }\rev{the }Euclidean squared norm \rev{of the error}, which is a weighted sum of squares of the minimization objectives in position and orientation:
\begin{align}
    ||\boldsymbol{e}||^2 &= ||\boldsymbol{e}(\boldsymbol{p}_{\mathrm{des}}, \boldsymbol{p}(t))||^2 \nonumber \\
    &= k_{\mathrm{pos}}^2 ||\boldsymbol{x}_{\mathrm{des}} - \boldsymbol{x}(t)||^2 + k_{\mathrm{ori}}^2 \theta_{\mathrm{des}, t}^2.
\end{align}
The algorithm in \secref{sc:mainalg} is developed to respond to an externally-specified~$\boldsymbol{\mathcal{V}}_{\mathrm{des}}$, being considered successful if it makes it possible to operate the end-effector from any starting pose \new{to any reachable pose} (as long as there exists a connected, everywhere-reachable path between them). It is reasonable to assume that~$\boldsymbol{\mathcal{V}}_{\mathrm{des}}$ is generated in such a manner that it is proportional to the error~$\boldsymbol{e}$ between the current and next desired poses. It is based on this proportional relationship that \secref{sc:contproof} and \secref{sc:discrete} approach the \new{local} stability analysis in terms of the problem of  reaching~$\boldsymbol{p}_{\mathrm{des}}$.

\section{Algorithm}
\label{sc:mainalg}
The most basic version of the algorithm is presented in this section, together with a proof of stability in \secref{sc:contproof}. \secref{sc:discrete} elaborates upon the considerations for a discrete-time implementation, and \secref{sc:mods} discusses important modifications for practical use. 

In order to achieve the following objectives,\begin{enumerate}
    \item yield the same result as the standard Jacobian inversion \rev{or pseudoinversion} when far from a singularity;
    \item near and at a singular \rev{configuration}\del{ pose}, only attempt task-space motions that respect the \new{altered kinematics, i.e., account for the extent and directions in which mobility is reduced}. \rev{Allow }\del{Ensure that} the singular \rev{pose }\del{ point} itself\rev{, or any unreachable pose, to also} \del{can} be set as a target pose; \new{and}
    \item \new{allow} small perturbations from the singular pose \new{to cause} the manipulator to retreat from the singularity to a new desired pose\new{,}
\end{enumerate} 
the algorithm relies \new{up}on three components: the Safety Jacobian~($\boldsymbol{J}_s$) (\secref{sc:safety_jacobian}), the Projection Jacobian~($\boldsymbol{J}_p$) (\secref{sc:projection_jacobian}) and Singular Projections~($\boldsymbol{\Phi}$,~$\tilde{\boldsymbol{U}}$) (\secref{sc:singular_projections}). 

\new{\textbf{Summary}: }Briefly, the purpose of~$\boldsymbol{J}_s$ is to remain as a non-singular approximation of the true~$\boldsymbol{J}$, so that joint speeds may be computed corresponding to an appropriately-modified~$\boldsymbol{\mathcal{V}}_{\mathrm{des}}$. This modification relies on separating the components of the requested~$\boldsymbol{\mathcal{V}}_{\mathrm{des}}$ that should not be affected by the singularity (identified using~$\boldsymbol{J}_p$, and preserved), from those that demand unbounded joint speeds (scaled down using~$\boldsymbol{\Phi}$,~$\tilde{\boldsymbol{U}}$). A\new{n implementation} of the algorithm is given in \figref{fig:pseudocode}. In the following, the subscript on~$\boldsymbol{\mathcal{V}}_{\mathrm{des}}$ is dropped for clarity.
\begin{figure}[!t] 
\centering
{\scriptsize
\begin{algorithmic} 
\algrule
\REQUIRE $\gamma, \boldsymbol{J}_{m \times n}$
\ENSURE $\boldsymbol{J}^+_{\mathrm{parse}}$
\algrule
\vspace{0.5em}
\STATE $\boldsymbol{U}, \new{\boldsymbol{\Sigma}}, \boldsymbol{V} \gets \textrm{SVD}(\boldsymbol{J})$ 
\STATE $\new{\boldsymbol{W}} \gets \new{\boldsymbol{\Sigma}}^\top$\COMMENT{\texttt{//set diagonal matrix dimensions}}
\STATE $b \gets \gamma \max(\new{\boldsymbol{\Sigma}})$\COMMENT{\texttt{//compute}~$\gamma\sigma_{\mathrm{max}}$~\texttt{threshold once}}
\FOR {$i \in \{1, \dots, m\}$} 
        \IF {$\new{W}_i < b$}
            \STATE $\new{W}_i \gets \frac{\new{W}_i}{b^2}$ \ICOMMENT{\texttt{//below threshold, set to}~$\frac{\sigma_i}{(\gamma \sigma_{\mathrm{max}})^2}$}
        \ELSE 
            \STATE $\new{W}_i \gets \frac{1}{\new{W}_i}$ \ICOMMENT{\texttt{//for other directions, set to}~$\frac{1}{\sigma_i}$}
        \ENDIF
\ENDFOR \COMMENT{\texttt{//keep off-diagonal elements zero}}
\STATE $\boldsymbol{J}_{\mathrm{parse}}^+ \gets \boldsymbol{V} \new{\boldsymbol{W}} \boldsymbol{U}^\top$
\RETURN $\boldsymbol{J}_{\mathrm{parse}}^+$
\vspace{0.5em}
\algrule
\end{algorithmic}
}
\caption{\new{Pseudocode for real-time implementation (\eqref{eqn:vwu},~\eqref{eqn:wdef}).}} 
\label{fig:pseudocode} 
\vspace{-2em}
\end{figure}

\subsection{Safety Jacobian}
\label{sc:safety_jacobian}
The Safety Jacobian~$\boldsymbol{J}_s$ is calculated by performing \gls{SVD} on the original geometric Jacobian matrix~$\boldsymbol{J}$ and inspecting every singular value in~$\boldsymbol{\Sigma}$ from~\eqref{eqn:J_svd}. A threshold value $\gamma \in (0,1]$ is selected to define the neighborhood of a singularity as~\footnote{\new{In a serial manipulator,~$\sigma_{\mathrm{max}}$ is always non-zero (\secref{sc:gamma}).}}:
\begin{equation}
    \new{\frac{\sigma_{\mathrm{min}}}{\sigma_{\mathrm{max}}} = \frac{1}{\kappa_{\mathrm{cn}}} < \gamma}.
\end{equation}
The term ``singular direction" hereinafter refers to the left singular vector associated with any~$\sigma_i$ satisfying ${\sigma_i < \gamma \sigma_{\mathrm{max}}}$\new{, denoted as~$\sigma_i^*$. That is, if the singular values are indexed in the conventional decreasing order, and~$\gamma \sigma_{\mathrm{max}}$ exceeds~$k$ of them,
\begin{align}
\sigma_1 \geq \dots \sigma_{m-k} \geq \gamma\sigma_{\mathrm{max}} \geq \sigma_{m - k +1}^* \geq \dots \sigma_m^*. \nonumber
\end{align}}
Replacing all such~$\sigma_i^{\new{*}}$ in~$\boldsymbol{\Sigma}$, a new matrix~$\boldsymbol{\Sigma}_s$ is constructed of the form given in \eqref{eqn:expanded_singular_matrix}:
\begin{equation}
    \Sigma_{s, \new{i}} = \begin{cases}
        \sigma_i, & \text{if } \sigma_i \geq \gamma \sigma_{\mathrm{max}} \\
        \gamma \sigma_{\mathrm{max}}, & \text{if } \sigma_i < \gamma \sigma_{\mathrm{max}}.
    \end{cases}
\end{equation}
With this new $\boldsymbol{\Sigma}_s$, the matrix $\boldsymbol{J}_s$ is then composed as: 
\begin{equation}
    \boldsymbol{J}_{s, m \times n} = \boldsymbol{U}_{m \times m} \boldsymbol{\Sigma}_{s, m \times n} \boldsymbol{V}^\top_{n \times n}
    \label{eqn:safety_jacobian}
\end{equation}

\subsection{Projection Jacobian}
\label{sc:projection_jacobian}
The component~$\boldsymbol{\mathcal{V}}_p$ of~$\boldsymbol{\mathcal{V}}_{\mathrm{des}}$ that is composed of the non-singular directions alone can be obtained by first inverting a Jacobian matrix which has singular values in only those directions, to obtain the necessary joint-space velocity~$\dot{\boldsymbol{q}}_p$, and then using the same Jacobian matrix to project back into the task space: 
\begin{equation}
    \boldsymbol{\mathcal{V}}_p = \boldsymbol{J}_p \dot{\boldsymbol{q}}_{\rev{p}} = \boldsymbol{J}_p \boldsymbol{J}_p^{+} \boldsymbol{\mathcal{V}}
    \label{eqn:Jp_derivation_explained}
\end{equation}
To obtain $\boldsymbol{J}_p$, the columns of $\boldsymbol{U}$ and rows of $\boldsymbol{\Sigma}$ which correspond to singular directions are dropped, such that for~$k$ singular directions:
\begin{equation}
    \boldsymbol{J}_{p, m \times n} = \boldsymbol{U}_{\rev{p,} m \times (m-k)} \boldsymbol{\Sigma}_{p,(m-k) \times n} \boldsymbol{V}^\top_{n \times n}
    \label{eqn:Jp_derivation}
\end{equation}
As~${\textrm{rank}(\boldsymbol{J}_p) < m}$, a true right-inverse does not exist and the alternative definition of pseudoinverse for singular matrices (Section~\ref{sc:singps}) is used, i.e.,~${\boldsymbol{J}_p \boldsymbol{J}_p^{+} \ne \boldsymbol{I}_{m \times m}}$. Rather,~${\boldsymbol{J}_p \boldsymbol{J}_p^{+} = \boldsymbol{U}_p \boldsymbol{U}^\top_p}$ is an~${m \times m}$ projection matrix onto the non-singular directions in the task space\footnote{\new{Equivalently, the singular values for singular directions can set to zero, instead of dropping the corresponding rows and columns of~$\boldsymbol{\Sigma},\boldsymbol{U}$. This results in the same~$m \times m$ projection matrix as~$\boldsymbol{J}_p \boldsymbol{J}_p^{+}$.}}. That is,~$\boldsymbol{\mathcal{V}}_p$ may be more efficiently computed as:
\begin{equation}
    \boldsymbol{\mathcal{V}}_p = \boldsymbol{U}_p \boldsymbol{U}_p^\top \boldsymbol{\mathcal{V}}\label{eqn:Jp_derivation_explained_more}
\end{equation}

\subsection{Singular Projections}
\label{sc:singular_projections}
The matrix~$\tilde{\boldsymbol{U}}_{m \times k}$ is constructed from~$\boldsymbol{U}_{m \times m}$ by retaining only the~$k$ columns corresponding to the singular directions. The commanded vector component in the \rev{singular} direction~$\tilde{\boldsymbol{u}}_i$ \del{of $\tilde{\boldsymbol{U}}$} is written as $\tilde{\boldsymbol{\mathcal{V}}} = (\tilde{\boldsymbol{u}}_i^\top \boldsymbol{\mathcal{V}})\tilde{\boldsymbol{u}}_i$. These components $\tilde{\boldsymbol{\mathcal{V}}}$ are then scaled by the ratio of the singular value for that direction to the threshold minimum singular value:
\begin{align}
    \tilde{\boldsymbol{\mathcal{V}}}' &= \left( \frac{\sigma_i}{\gamma \sigma_{\mathrm{max}}} \right) \tilde{\boldsymbol{\mathcal{V}}} \nonumber \\
    &= \left( \frac{\sigma_i}{\gamma \sigma_{\mathrm{max}}} \right) (\tilde{\boldsymbol{u}}_i^\top \boldsymbol{\mathcal{V}})\tilde{\boldsymbol{u}}_i
    \label{eqn:singular_direction_projection_vector_form}
\end{align}
The diagonal matrix~$\boldsymbol{\Phi}$ is defined to contain the $k$ singular value ratios:
\begin{equation}
    \boldsymbol{\Phi}_{k \times k} =  \begin{bmatrix}
        \frac{\sigma^*_{\new{m-k+1}}}{\gamma \sigma_{\mathrm{max}}} & \ldots & 0 \\ 
        \vdots & \ddots & \vdots \\
        0 & \ldots & \frac{\sigma^*_{\new{m}}}{\gamma \sigma_{\mathrm{max}}}
    \end{bmatrix}
    \nonumber
\end{equation}
Combining this with \eqref{eqn:singular_direction_projection_vector_form} and writing in matrix form produces:
\begin{equation}
    \tilde{\boldsymbol{\mathcal{V}}}_{m\times 1}' = \tilde{\boldsymbol{U}}_{m \times k} \boldsymbol{\Phi}_{k \times k} \tilde{\boldsymbol{U}}^\top_{k\times m}  \boldsymbol{\mathcal{V}}_{m \times 1} 
    \nonumber
\end{equation}

\subsection{J-PARSE Algorithm}
With these definitions, the full J-PARSE algorithm can be constructed: 
\begin{align}
    \dot{\boldsymbol{q}}_{\mathrm{des}} &= \boldsymbol{J}_s^{+} (\boldsymbol{\mathcal{V}}_p + \boldsymbol{\mathcal{V}}'_{\rev{s}}) \nonumber \\
    &= \underbrace{\boldsymbol{J}_s^{+} \left( \rev{\boldsymbol{U}}_p \rev{\boldsymbol{U}}_p^\top + \tilde{\boldsymbol{U}} \boldsymbol{\Phi} \tilde{\boldsymbol{U}}^\top\right)}_{\boldsymbol{J}_{\mathrm{parse}}^+} \boldsymbol{\mathcal{V}} 
    \label{eqn:jparse_full}
\end{align}
Reconstituting~$\boldsymbol{U}$ from the appropriately-ordered columns of~$\boldsymbol{U}_p$ and~$\tilde{\boldsymbol{U}}$, a more compact form is given by:
\begin{align}
    \dot{\boldsymbol{q}}_{\mathrm{des}}
    &= \boldsymbol{J}_s^{+} \left(\boldsymbol{U}\boldsymbol{S} \boldsymbol{U}^\top \right) \boldsymbol{\mathcal{V}},  \label{eqn:jparse_simple}\\
    &= \boldsymbol{V} \boldsymbol{\Sigma}_s^{+} \boldsymbol{S} \boldsymbol{U}^\top  \boldsymbol{\mathcal{V}},
    \label{eqn:jparse_simples}
\end{align}
where~$\boldsymbol{S}_{m \times m}$ contains
\begin{align}
    S_{ij} = \begin{cases}
                        0 & \text{if } i \neq j \\ 
                        1 & \text{if } i = j \text{ and } \sigma_i \geq \gamma \sigma_{\mathrm{max}} \\
                         \frac{\sigma_i}{\gamma \sigma_{\mathrm{max}}} & \text{if } i = j \text{ and } \sigma_i < \gamma \sigma_{\mathrm{max}}
                    \end{cases}
    \label{eq:sij}
\end{align}
The following features are evident from the above:
\begin{enumerate}
    \item In non-singular configurations,~$\boldsymbol{\Sigma}_s^+ = \boldsymbol{\Sigma}^+$ and~$\boldsymbol{S} = \boldsymbol{I}$, so the regular pseudoinverse is used.
    \item When exactly at a singular configuration (even if the goal pose is outside the reachable workspace), no motion is requested in the singular direction(s), as~${\sigma_i = 0 \implies \Phi_i = 0}$. In the immediate neighborhood of the singularity, the requested motion in these directions remains small, as~$\sigma_i << \gamma \sigma_{\mathrm{max}}$.
    \item The entries of~$\boldsymbol{\Phi}$, and therefore, the joint speeds~$\dot{\boldsymbol{q}}$, are continuous across the boundary of the singular region.
\end{enumerate}
\new{For more efficient implementation, given in \figref{fig:pseudocode}, the expression in~\eqref{eqn:jparse_simples} is rewritten as:
\begin{align}
    \dot{\boldsymbol{q}}_{\mathrm{des}}
    &= \boldsymbol{V} \boldsymbol{W} \boldsymbol{U}^\top  \boldsymbol{\mathcal{V}}, \label{eqn:vwu}
\end{align}
where~$\boldsymbol{W}$ has the same dimensions as~$\boldsymbol{\Sigma}^+$ or~$\boldsymbol{\Sigma}^+_s$, but with diagonal elements~$W_i = \boldsymbol{\Sigma}^+_{s,i} S_i$.}

\subsection{Note on Escaping Singularities}
\label{sc:esc}
\new{\textit{Exactly singular} configurations, as a subset of the configuration space, belong to a set of measure zero\footnote{\new{The function~$\det{(\boldsymbol{J})}$ is a real analytic function that is not identically zero on the configuration space, \textit{therefore} its zero set~$\{\boldsymbol{q}: \det{(\boldsymbol{J})} = 0\}$ (the set of singular configurations) has measure zero (as shown in~\cite{Mityagin2015TheZS}).}}. Therefore, every configuration with~${\sigma_{\mathrm{min}} = 0}$ has infinitely many neighbors where~${\sigma_{\mathrm{min}} > 0}$. }In practice, \new{this means that~$\sigma_{\mathrm{min}}$ is \textit{almost never}\footnote{\new{The mathematical definition of ``almost never'' assigns a \textit{probability} of zero to a \textit{possible} event, if it only occurs on a set of measure zero in a probability space~\cite{chan2013measure, weisstein2024almostsurely}. If measurement noise is modeled as a random continuous variable, the robot can ``almost never'' be measured to be exactly at a singular configuration. Although neither encoder outputs nor singular values are truly continuous in reality, in practice, the variation in~$\sigma_{\mathrm{min}}$ due to sensor noise is expected to be several orders of magnitude higher than the machine precision, meaning that the sensor samples effectively continuously from a neighborhood of the singularity. This makes it ``almost certain'' that true zero is not sampled.}} exactly~$0$. Instead, the singular values arbitrarily close to singularity (and the corresponding values of~$S_i$) are still very small positive numbers\footnote{\new{SVD algorithms do not output negative singular values. They flip the singular vectors if necessary, to ensure that singular values are non-negative~(\cite{golub1970singular}, p.~413). Thus, even floating point errors cannot make singular values fall below zero.}}.} This permits small motions that eventually enable the manipulator to escape singular regions, \new{as the speed increases gradually along with its emergence from the singular region.}

\new{It is instructive to note that \textit{if} the smallest singular value were artificially set to be truly zero, up to the full precision used by the system, then any velocity commanded along the singular direction(s) would in fact be scaled down to zero. This behavior would be in agreement with, and in continuity with, the increasingly-unavailable mobility.} 

\new{While~$S_i$ is deliberately allowed to become arbitrarily small in this formulation, the inversion itself always remains stable, as it is ensured in the construction of~$\boldsymbol{J}_s$ that~${\Sigma_{s, i} \geq \gamma \sigma_{\mathrm{max}}}$.} 
\begin{figure*}[!htbp] 
    \centering
    \includegraphics[width=0.7\linewidth]{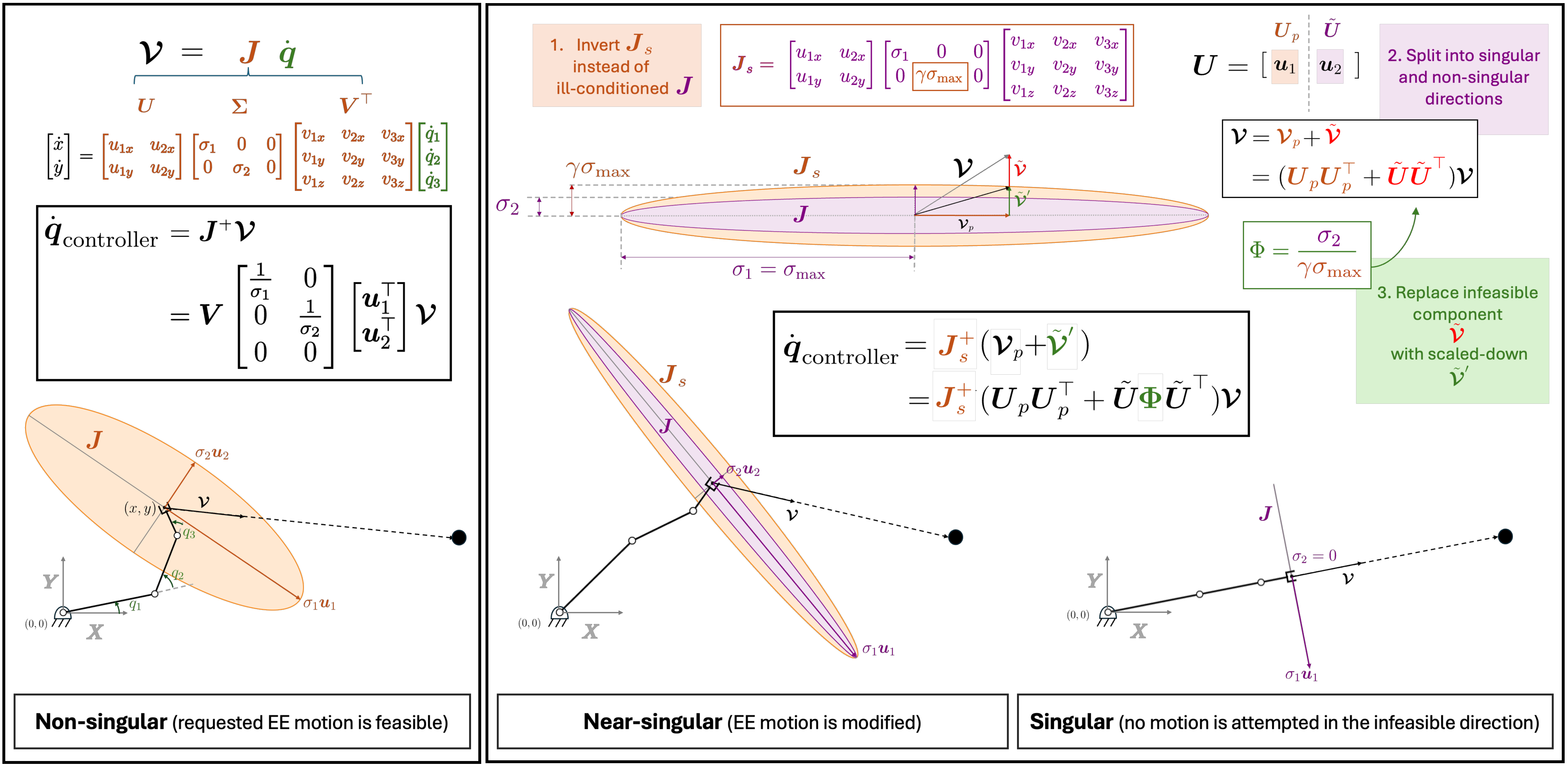} 
    \caption{A summary of the J-PARSE algorithm using the example of a planar~$3$-R manipulator. Here, the end-effector (EE) twist is simply linear velocity. When~$\boldsymbol{J}$ is well-conditioned, the regular pseudoinverse is used. When the ICN of~$\boldsymbol{J}$ falls below~$\gamma$, the safety Jacobian matrix~$\boldsymbol{J}_s$ (corresponding to a velocity ellipse with the same major axis as~$\boldsymbol{J}$, but aspect ratio~$\gamma$) and differentially-scaled projections of~$\boldsymbol{\mathcal{V}}$ are used to compute an appropriate set of joint speeds, as shown.}
    \label{fig:jparse_alg_summary}
    \vspace{-1em}
\end{figure*}

\section{Continuous-time stability analysis}
\label{sc:contproof}
As it is customary to first prove the stability of the algorithm in a continuous-time framework, the manipulator controlled with the proposed algorithm, setting~${\boldsymbol{\mathcal{V}} = \boldsymbol{e}(\boldsymbol{p}_{\mathrm{des}}, \boldsymbol{p})}$, is proved to be locally stable everywhere as follows. Without loss of generality,~${\boldsymbol{p}_{\mathrm{des}}}$ may be chosen as the reference for measuring position as well as orientation. Then, from~\eqref{eq:errordef}:
\begin{align}
    \boldsymbol{e}(\underbrace{\boldsymbol{p}_{\mathrm{des}}}_{\boldsymbol{0}}, \boldsymbol{p}) 
    = \new{-}\begin{bmatrix}
        k_{\mathrm{pos}} \boldsymbol{x} \\
        k_{\mathrm{ori}} \theta \hat{\boldsymbol{h}}
    \end{bmatrix} = \new{-}\boldsymbol{K} \begin{bmatrix}
        \boldsymbol{x} \\
        \theta \hat{\boldsymbol{h}}
    \end{bmatrix} = \new{-}\boldsymbol{K} \boldsymbol{p},
\end{align}
where the diagonal positive-definite matrix~$\boldsymbol{K}$ functions as a proportional gain (discussed in further detail in \ref{sc:discrete}). \new{It may be replaced, as needed, by a fully diagonal decoupled gain matrix~${\boldsymbol{K} = \mathrm{diag}(k_x, k_y, k_z, k_{\omega x}, k_{\omega y}, k_{\omega z})}$, without affecting the analyses presented in the remainder of this paper.} 

The task-space pose~$\boldsymbol{p}$ is a function of the joint space configuration~$\boldsymbol{q}$ (but not vice versa, \new{as IK solutions are non-unique}). The state of the system is therefore described by~$\boldsymbol{q}$, which also determines the Jacobian matrix~$\boldsymbol{J}(\boldsymbol{q})$. The system evolves as (from \eqref{eqn:vwu}):
\begin{align}
    \dot{\boldsymbol{q}} = \boldsymbol{f}(\boldsymbol{q}) = \boldsymbol{V}\boldsymbol{W}\boldsymbol{U}^\top (\new{-}\boldsymbol{K}\boldsymbol{p}),
    \label{eq:qoft}
\end{align}
where~$\boldsymbol{V},~\boldsymbol{W},~\boldsymbol{U},~\boldsymbol{p}$ are all functions of~$\boldsymbol{q}$. \new{The function~$\boldsymbol{f}(\boldsymbol{q})$ is time-independent (a function of the state alone). Therefore, }\emph{provided~$\boldsymbol{f}(\boldsymbol{q})$ is locally Lipschitz, local stability can be analyzed using Lyapunov's second method~\cite{liberzon2003switching}.} That is, the system is stable if there exists a continuous function~\new{$V(\boldsymbol{q})$ (a Lyapunov function)} that is nonincreasing with time when~$\boldsymbol{q}(t)$ satisfies \eqref{eq:qoft}. \new{The system} is locally asymptotically stable if \new{a~$V(\boldsymbol{q})$ exists that} is strictly decreasing. 

\subsection{Preconditions for Lyapunov stability analysis}
\new{The matrix}~$\boldsymbol{W}_{n \times m}$ contains
\begin{align}
    W_{ij} = \begin{cases}
                        0 & \text{if } i \neq j \\ 
                        \frac{1}{\sigma_i} & \text{if } i = j \text{ and } \sigma_i \geq \gamma \sigma_{\mathrm{max}} \\
                         \frac{\sigma_i}{(\gamma \sigma_{\mathrm{max}})^2} & \text{if } i = j \text{ and } \sigma_i < \gamma \sigma_{\mathrm{max}}
                    \end{cases}. \label{eqn:wdef}
\end{align}
Sums and products of locally Lipschitz functions are also themselves locally Lipschitz. Therefore, it is sufficient to show that the elements of~${\boldsymbol{V}\boldsymbol{W}\boldsymbol{U}^\top,~\boldsymbol{K}}$, and~${\boldsymbol{p}}$ are continuous and have bounded derivatives with respect to~$\boldsymbol{q}$ in the neighborhood under consideration. 

The gain matrix~$\boldsymbol{K}$ is independent of~$\boldsymbol{q}$ in the nominal case (or if it is varied with the state, can be chosen to be varied in a smooth manner with a bounded derivative). For a sufficiently small~$\boldsymbol{p}$, the derivatives of~$\boldsymbol{p}$ w.r.t.~$\boldsymbol{q}$ are the twists\footnote{See also~\eqref{eqn:e_p_diff_vel}-\eqref{eq:smalltwist} in~\secref{sc:vdot}.} given by the columns of~$\boldsymbol{J}$, elements of which are polynomial expressions involving link geometry parameters and algebraic or trigonometric functions of joint states (of prismatic or revolute joints respectively). Elements of~$\boldsymbol{J}$ are therefore continuous and bounded. This leaves only the preceding expression~$\boldsymbol{V}\boldsymbol{W}\boldsymbol{U}^\top$.

In a typical implementation of SVD,~$\boldsymbol{U}$ and~$\boldsymbol{V}$ need not vary continuously with~$\boldsymbol{q}$. Discontinuities may arise in the following ways:\\
\textbf{1) Sign changes}: One or more singular vectors may flip direction. However, whenever a left singular vector (column of~$\boldsymbol{U}$) flips direction, it is accompanied by a flip in the direction of the right singular vector (row of~$\boldsymbol{V}^\top$) corresponding to the same singular value. Therefore, such discontinuities do not appear in the product~$\boldsymbol{V}\boldsymbol{W}\boldsymbol{U}^\top$.\\
\textbf{2) Multiplicity}: When multiple singular values become equal, the corresponding singular vectors are not unique. For example, a discontinuity is possible when a previously-unique singular value attains a multiplicity of two, as the corresponding columns may be replaced by any two linearly independent orthogonal vectors in the space spanned by those columns. Again, in such a state, if \new{a pair (or set) of} columns \new{of}~$\boldsymbol{U}$~\new{is replaced by a transformed set of vectors}, the\new{n a corresponding set of} rows of~$\boldsymbol{V}^\top$ \new{is also transformed,} such that the product~$\boldsymbol{V}\boldsymbol{W}\boldsymbol{U}^\top$ remains unaffected. 

Thus, since~$\boldsymbol{U}$ and~$\boldsymbol{V}$ never appear in isolation but only in the form~$\boldsymbol{V}\boldsymbol{W}\boldsymbol{U}^\top$, discontinuities in the SVD do not appear in~$\dot{\boldsymbol{q}}$. Having ruled out effects of such discontinuities, it is noted that the matrices themselves are bounded (both are orthogonal), as are their derivatives w.r.t~$\boldsymbol{q}$~\cite{jacsvd}.

Finally, the elements of~$\boldsymbol{W}$ are examined for the Lipschitz property. Off-diagonal elements are~$0$. Diagonal elements depend upon singular values, which are indirectly dependent \new{up}on~$\boldsymbol{q}$. Since a closed-form relationship is not available, the singular values are related to~$\boldsymbol{q}$ via the elements of~$\boldsymbol{J}$ itself:
\begin{align}
    \frac{\partial\sigma_i}{\partial{q_k}} &= \sum_{l,h} \frac{\partial\sigma_i}{\partial{J_{lh}}} \frac{\partial{J_{lh}}}{\partial{q_k}} \\
    &= \sum_{l,h} U_{li} V_{hi} \frac{\partial{J_{lh}}}{\partial{q_k}},
    \label{eq:sbyq}
\end{align}
based on the derivations in~\cite{jacsvd}. As mentioned above, the elements of~$\boldsymbol{U},\boldsymbol{V}$ are bounded since they are orthogonal matrices, and the elements of~$\boldsymbol{J}$ are polynomial expressions involving the joint parameters and their trigonometric functions, the derivatives of which are therefore guaranteed to be bounded. Hence, the expressions in~\eqref{eq:sbyq} are bounded. Now, the variation of~$W_{i\new{j}}$ with~$q_\new{k}$ is given by:
\begin{align}
    \frac{\partial{W_{ij}}}{\partial q_k} = \begin{cases}
                        0 & \text{if } i \neq j \\ 
                        -\frac{1}{\sigma_i^2} \frac{\partial\sigma_i}{\partial{q_\new{k}}} & \text{if } i = j \text{,} \sigma_i \geq \gamma \sigma_{\mathrm{max}} \\
                         \frac{1}{\gamma^2 \sigma_{\mathrm{max}}^2} \frac{\partial\sigma_i}{\partial{q_\new{k}}} - 2 \frac{\sigma_i}{\gamma \sigma_{\mathrm{max}}^3} \frac{\partial\sigma_{\mathrm{max}}}{\partial{q_\new{k}}} & \text{if } i = j \text{,} \sigma_i < \gamma \sigma_{\mathrm{max}}
                    \end{cases},
                    \label{eq:wbyq}
\end{align}
where~${\sigma_{\mathrm{max}} = \sigma_\new{1}}$, by choice of ordering in SVD, and also varies according to~\eqref{eq:sbyq}. Using~\eqref{eq:sbyq} in~\eqref{eq:wbyq}, it is obvious that~$W_i$ has bounded derivatives wherever the derivative is defined, i.e., everywhere except for the countable set of transition points where~${\sigma_i}$~\new{crosses~}${\gamma \sigma_{\mathrm{max}}}$. At these transition points, derivatives on both sides are bounded. Therefore,~$\boldsymbol{f}$ is locally Lipschitz everywhere\footnote{A function that is locally integrable and weakly differentiable with bounded derivatives on an open interval is Lipschitz on the corresponding closed interval (Theorem~3.53 in~\cite{hunternotes}); here,~$\boldsymbol{f}$ is locally integrable since it is continuous, and is weakly differentiable since the transition points belong to a set of measure zero.}. 

\subsection{Lyapunov stability analysis}
\label{sc:vdot}
To proceed with stability analysis, a candidate Lyapunov function is defined as follows.
\begin{equation}
    V\rev{(\ins{\boldsymbol{q}})} = \frac{1}{2} \boldsymbol{e}^\top \rev{\boldsymbol{K}^{-1}} \boldsymbol{e} \rev{= \frac{1}{2} \boldsymbol{p}^\top \boldsymbol{K} \boldsymbol{p}}. 
    \label{eqn:vel_cntrl_lyapunov}
\end{equation}
With~$V$ thus being defined as a quadratic form on the task-space error, controller stability implies non-increasing distance from the goal, and asymptotic stability implies approaching the goal without increasing distance \emph{until the goal is achieved}. We \new{assume} that~$\boldsymbol{p}_{\mathrm{des}}$~\new{(defined as the origin)} is reachable, that is, there exists at least one~$\boldsymbol{q}_0$ such that~$\boldsymbol{p}(\boldsymbol{q}_0) = \boldsymbol{p}_{\mathrm{des}}$. While~$\boldsymbol{q}_0$ is not unique in general, it is always possible to define a neighborhood of the current~$\boldsymbol{q}$ within which~$\boldsymbol{q}_0$ is unique\footnote{\new{As discussed in \secref{sc:singps}, IK solutions coincide only at singular configurations. Everywhere else, IK is bijective. If~$\boldsymbol{q}$ is singular, then there may be multiple nearby~$\boldsymbol{q}_0$ satisfying~${\boldsymbol{p}(\boldsymbol{q}_0) = \boldsymbol{p}_{\mathrm{des}}}$, but there is still \textit{at least} one acceptable~$\boldsymbol{q}_0$ and the neighborhood may be defined asymmetrically in such a manner as to exclude all others.} The problem of \new{selecting the} branch\new{/posture} selection when exiting a \new{singular} configuration is addressed separately in \secref{sc:branches}.}. Therefore, the conditions~${V(\ins{\boldsymbol{q}_0}\neq0) > 0}$ and~${V(\ins{\boldsymbol{q}_0}=0)=0}$ are satisfied. The \ins{time} derivative of $V$ is
\begin{equation}
    \dot{V} = \frac{\partial V }{\partial \rev{\boldsymbol{p}}} \frac{\textrm{d}\rev{\boldsymbol{p}}}{\textrm{d}t} = \rev{\boldsymbol{p}}^\top \rev{\boldsymbol{K}} \rev{\dot{\boldsymbol{p}}}.
    \label{eqn:lyp_lie_deriv}
\end{equation}
The condition for \new{local }stability is~${\dot{V} \leq 0}$, and for \new{local} asymptotic stability, the strict inequality~${\dot{V} < 0}$. 
Differentiating~$\boldsymbol{p}$ with respect to time, 
\begin{equation}
    \rev{\dot{\boldsymbol{p}}} = \frac{\textrm{d}\boldsymbol{p}}{\textrm{d}t} =  \begin{bmatrix}
    \dot{\boldsymbol{x}} \\
    \dot{\theta} \hat{\boldsymbol{h}} + \theta \dot{\hat{\boldsymbol{h}}}
    \end{bmatrix} = \begin{bmatrix}
    \boldsymbol{v} \\
    \boldsymbol{\omega} + \theta \dot{\hat{\boldsymbol{h}}}
    \end{bmatrix} 
    \label{eqn:e_p_diff_vel}
\end{equation}
If the orientations of~$\boldsymbol{p}_{\mathrm{des}}$ and~$\boldsymbol{p}$ are very close to each other, then~\ins{$\theta$ is small, and }~$\frac{\textrm{d}}{\textrm{d}t}\left(\theta \hat{\boldsymbol{h}}\right) \approx \dot{\theta} \hat{\boldsymbol{h}}$. In this case,\rev{~$\dot{\boldsymbol{p}}$ approaches the twist~$\boldsymbol{\mathcal{V}}$}:
\begin{align}
    \rev{\dot{\boldsymbol{p}}} =  \begin{bmatrix}
    \boldsymbol{v} \\
    \boldsymbol{\omega} 
    \end{bmatrix} = \boldsymbol{J}\dot{\boldsymbol{q}}
    \label{eq:smalltwist}
\end{align}
To achieve the desired motion, substitute \eqref{eqn:jparse_simple} into \eqref{eqn:e_p_diff_vel} for $\dot{\boldsymbol{q}}$, and\rev{~$\boldsymbol{\mathcal{V}} = - \boldsymbol{K}\boldsymbol{p}$}:
\begin{align}
    \dot{\boldsymbol{p}} =  \boldsymbol{J} \dot{\boldsymbol{q}} &=  \boldsymbol{J} \boldsymbol{J}_s^{+} \boldsymbol{U}\boldsymbol{S} \boldsymbol{U}^\top \boldsymbol{\mathcal{V}} \\
    &=   \boldsymbol{U} \boldsymbol{\Sigma} \boldsymbol{V}^\top \boldsymbol{V} \boldsymbol{\Sigma}_s^{+} \boldsymbol{S} \boldsymbol{U}^\top \boldsymbol{\mathcal{V}} \\
    &=   \boldsymbol{U} \boldsymbol{\Sigma} \boldsymbol{\Sigma}_s^{+} \boldsymbol{S} \boldsymbol{U}^\top \boldsymbol{\mathcal{V}} \\
    &= -  \boldsymbol{U} \boldsymbol{\Sigma} \boldsymbol{\Sigma}_s^{+} \boldsymbol{S} \boldsymbol{U}^\top \boldsymbol{K} \boldsymbol{p}
    \label{eqn:expanded_error_deriv_jparse_stability}
\end{align}
Substituting \eqref{eqn:expanded_error_deriv_jparse_stability}, into \eqref{eqn:lyp_lie_deriv} \del{then into (33) }yields the following condition for stability:
\begin{align}
        \rev{- \boldsymbol{p}^\top \boldsymbol{K} \boldsymbol{U} \left( \boldsymbol{\Sigma} \boldsymbol{\Sigma}_s^{+} \boldsymbol{S} \right) \boldsymbol{U}^\top \boldsymbol{K} \boldsymbol{p} \leq 0}
    \label{eqn:stability_initial_sub_vel}
\end{align}
Since~$\boldsymbol{U}$ is orthogonal and~$\boldsymbol{K}$ is diagonal and invertible, for every~$\boldsymbol{p}$, there is a unique~${\boldsymbol{p}^\prime = \boldsymbol{U}^\top \boldsymbol{K} \boldsymbol{p}}$ and the condition reduces to:
\begin{align}
        - {\boldsymbol{p}^\prime}^\top \underbrace{\left( \boldsymbol{\Sigma} \boldsymbol{\Sigma}_s^{+} \boldsymbol{S} \right)}_{\boldsymbol{Q}} \boldsymbol{p}^\prime \leq 0,
        \label{eqn:pprime}
\end{align}
where~$\boldsymbol{Q}$ is diagonal:
\begin{align}
    Q_{ij} = \begin{cases}
                        0 & \text{if } i \neq j \\
                        1 & \text{if } i = j \text{ and } \sigma_i \geq \gamma \sigma_{\mathrm{max}} \\
                         \frac{\sigma_i^2}{\gamma^2 \sigma_{\mathrm{max}}^2} & \text{if } i=j \text{ and } \sigma_i < \gamma \sigma_{\mathrm{max}}
    \end{cases}.
    \label{eqn:Qdef}
\end{align}
As the diagonal elements of~$\boldsymbol{Q}$ are non-negative, the local stability condition~\eqref{eqn:stability_initial_sub_vel} holds everywhere \new{(whether away from, near, or at singularities). Furthermore, the diagonal element~${\sigma_i > 0}$ at all non-singular configurations, including in the neighborhoods of singularities. Therefore, local asymptotic stability is guaranteed for all~$\boldsymbol{p}$ corresponding to non-singular configurations, \textit{including} starting and target configurations arbitrarily close to singular ones, as long as they are not \textit{exactly} singular. The case where the manipulator starts at a singular configuration is addressed in \secref{sc:esc}.}
\section{Discrete-time considerations}
\label{sc:discrete}
While the continuous-time stability analysis provides a satisfactory initial understanding of the behavior of the controller, such an analysis is not sufficient in itself, as~(a)~real-world implementation of digital control necessarily involves a finite time-step, and~(b)~the continuous-time analysis gives no insight into the considerations relevant to the selection of parameters. 

\subsection{Discrete-time local stability analysis}
\label{sc:discproof}
The discrete-time stability analysis follows a procedure outlined previously in the literature~\cite{das1988}. For a time step~$\Delta t$, the system dynamics can be approximated as a Taylor series: 
\begin{align}
    \boldsymbol{p}(t+\Delta t) = \boldsymbol{p}(t) + \Delta t ~\dot{\boldsymbol{p}}(t) + \frac{1}{2} (\Delta t)^2~\ddot{\boldsymbol{p}}(t) + \mathcal{O}((\Delta t)^3)) 
\end{align}
The second- and higher-order terms are dropped, which is permissible if~${||\dot{\boldsymbol{q}}||}$ is ``small"~\cite{das1988} (i.e.,~$\gamma$ is sufficiently large, see \secref{sc:gamma}). Using the same Lyapunov function as in the continuous-time analysis, the stability condition is (writing~$\boldsymbol{p}(t)$ simply as~$\boldsymbol{p}$ for clarity):
\begin{align}
    V(t+1) - V(t) &\leq 0  \nonumber \\
    \equiv (\boldsymbol{p} + \Delta t~\dot{\boldsymbol{p}})^\top ~\boldsymbol{K}(\boldsymbol{p} + \Delta t~\dot{\boldsymbol{p}}) - \boldsymbol{p}^\top~\boldsymbol{K}\boldsymbol{p} &\leq 0  \nonumber \\
    \equiv 2 (\Delta t) \boldsymbol{p}^\top \boldsymbol{K}\dot{\boldsymbol{p}} + (\Delta t)^2 \dot{\boldsymbol{p}}^\top \boldsymbol{K} \dot{\boldsymbol{p}} &\leq 0
\end{align}
Using~${\dot{\boldsymbol{p}}(t) = \boldsymbol{J}(t) \dot{\boldsymbol{q}}(t)}$ and rewriting:
\begin{align}
    2 (\Delta t) \boldsymbol{p}^\top \boldsymbol{K} \boldsymbol{J} \dot{\boldsymbol{q}} + (\Delta t)^2 \dot{\boldsymbol{q}}^\top \boldsymbol{J}^\top \boldsymbol{K} \boldsymbol{J} \dot{\boldsymbol{q}} &\leq 0
\end{align}
Substituting the controller definition~${\dot{\boldsymbol{q}} = \boldsymbol{J}_s^+ \boldsymbol{U} \boldsymbol{S} \boldsymbol{U}^\top \boldsymbol{\mathcal{V}}}$ from \eqref{eqn:jparse_simple}, and~${\boldsymbol{\mathcal{V}} = - \boldsymbol{K}\boldsymbol{p}}$:
\begin{align}
    &- 2 (\Delta t) \boldsymbol{p}^\top \boldsymbol{K} \boldsymbol{J} \boldsymbol{J}_s^+ \boldsymbol{U} \boldsymbol{S} \boldsymbol{U}^\top \boldsymbol{K} \boldsymbol{p} \nonumber \\
    &+ (\Delta t)^2 \boldsymbol{p}^\top \boldsymbol{K} \boldsymbol{U} \boldsymbol{S} \boldsymbol{U}^\top {\boldsymbol{J}_s^+} ^\top \boldsymbol{J}^\top \boldsymbol{K} \boldsymbol{J} \boldsymbol{J}_s^+ \boldsymbol{U} \boldsymbol{S} \boldsymbol{U}^\top \boldsymbol{K}\boldsymbol{p} \leq 0
\end{align}
Using the expansions~${\boldsymbol{J} = \boldsymbol{U} \boldsymbol{\Sigma} \boldsymbol{V}^\top}$ and~${\boldsymbol{J}_s^+ = \boldsymbol{V} \boldsymbol{\Sigma}_s^+ \boldsymbol{U}^\top}$, 
\begin{align}
    &- 2 (\Delta t) \boldsymbol{p}^\top \boldsymbol{K} \boldsymbol{U} \boldsymbol{\Sigma} \boldsymbol{\Sigma}_s^+ \boldsymbol{S} \boldsymbol{U}^\top \boldsymbol{K} \boldsymbol{p} \nonumber \\
    & \quad + (\Delta t)^2 \boldsymbol{p}^\top \boldsymbol{K} \boldsymbol{U} \boldsymbol{S}  {\boldsymbol{\Sigma}_s^+}^\top \boldsymbol{\Sigma}^\top \boldsymbol{U}^\top
     \boldsymbol{K} \boldsymbol{U} \boldsymbol{\Sigma}  \boldsymbol{\Sigma}_s^+ \boldsymbol{S} \boldsymbol{U}^\top \boldsymbol{K}\boldsymbol{p} \nonumber \\
     &=
     - 2 (\Delta t) \boldsymbol{p}^\top \boldsymbol{K} \boldsymbol{U} \boldsymbol{Q} \boldsymbol{U}^\top \boldsymbol{K} \boldsymbol{p} \nonumber \\
     & \quad + (\Delta t)^2 \boldsymbol{p}^\top \boldsymbol{K} \boldsymbol{U} \boldsymbol{Q} \boldsymbol{U}^\top
     \boldsymbol{K} \boldsymbol{U} \boldsymbol{Q} \boldsymbol{U}^\top \boldsymbol{K}\boldsymbol{p} \nonumber \\
     &=
     - {\boldsymbol{p}^\prime}^\top \underbrace{\left(
      2 (\Delta t) \boldsymbol{Q} 
     - (\Delta t)^2 \boldsymbol{Q} \boldsymbol{U}^\top
     \boldsymbol{K} \boldsymbol{U} \boldsymbol{Q}  \right)}_{2 (\Delta t)\boldsymbol{\Theta}} \boldsymbol{Q} \boldsymbol{p}^\prime \leq 0,
\end{align}
where (as in \eqref{eqn:pprime})~$\boldsymbol{p}^\prime = \boldsymbol{U}^\top \boldsymbol{K} \boldsymbol{p}$.
The discrete-time system is therefore stable if~$\boldsymbol{\Theta}$ is positive semi-definite:
\begin{align}
    \boldsymbol{\Theta} = \boldsymbol{Q}
     - \frac{\Delta t}{2} \boldsymbol{Q}  \boldsymbol{U}^\top
     \boldsymbol{K} \boldsymbol{U} \boldsymbol{Q}.
\end{align}
Treating the above as a symmetric perturbation of the diagonal matrix~$\boldsymbol{Q}$, a sufficient condition~${k \Delta t \leq \frac{2}{\ins{m(m-1)+1}}}$ (where~$k$ is an upper bound on the entries of~$\boldsymbol{K}$) is derived in \appref{app:suffcond}. 

In the special case~$\boldsymbol{K} = \ins{k} \boldsymbol{I}$, the matrix~$\boldsymbol{\Theta}$ is diagonal, and the condition is simplified:
\begin{align}
    Q_{i} - \ins{k} \frac{\Delta t}{2}{Q_{i}^2} \geq 0, \quad \forall i \in \{1, \dots, m\}, 
    \label{eq:conservativebound}
\end{align}
leading to~${k \Delta t \leq 2}$, in the corresponding units, for stable behavior, since~$Q_i \in [0, 1]$. This more permissive condition is preferred for most of the applications reported in~\secref{sc:results}.

\subsection{Selection of~\texorpdfstring{$\gamma$}{threshold}}
\label{sc:gamma}
The threshold ICN~$\gamma$, below which singularity handling is implemented, may be determined for a given manipulator by computing the ICN computationally while visualizing the manipulator at a range of configurations. Selection of~$\gamma$ may be used to derive an estimate of the highest joint speeds likely to be commanded. Alternatively, a desired maximum norm of~$\dot{\boldsymbol{q}}$ may be used to find a lower bound for the selection of~$\gamma$. Both approaches are based on the relationship:
\begin{align}
    ||\dot{\boldsymbol{q}}|| \leq \frac{1}{\gamma \sigma_{\mathrm{max}}} \new{\mathcal{V}}_{\mathrm{max}},
    \label{eq:gammsel}
\end{align}
where~$\new{\mathcal{V}}_{\mathrm{max}}$ is the maximum possible norm of~$\boldsymbol{\mathcal{V}}$ expected to be encountered as a request, since the greatest possible scaling in magnitude is by the largest singular value of~$\boldsymbol{J}^+_{\mathrm{parse}}$. 

Then, to satisfy~$||\dot{\boldsymbol{q}}|| \leq ||\dot{\boldsymbol{q}}||_{\mathrm{max}}$ over \emph{all} configurations, using~\eqref{eq:gammsel}, \new{it is sufficient to ensure that} the following \new{is} true everywhere:
\begin{align}
    &\frac{1}{\gamma \sigma_{\mathrm{max}}} \new{\mathcal{V}}_{\mathrm{max}} \leq ||\dot{\boldsymbol{q}}||_{\mathrm{max}}.
\end{align}
That is, for an expected~$\new{\mathcal{V}}_{\mathrm{max}}$ and desired~$||\dot{\boldsymbol{q}}||_{\mathrm{max}}$, it is sufficient to select~$\gamma$ satisfying
\begin{align}
    \gamma \geq \frac{\new{\mathcal{V}}_{\mathrm{max}}}{\mathrm{min}({\sigma_{\mathrm{max}})} ||\dot{\boldsymbol{q}}||_{\mathrm{max}}}.
\end{align}

In order to use this relationship, one may replace~$\mathrm{min}(\sigma_{\mathrm{max}})$ with a lower bound on~$\sigma_{\mathrm{max}}$ for the given manipulator, which is easily done based on its architecture. \new{In fact, usually~${\sigma_{\mathrm{max}} \geq 1}$, and it is often possible to identify a stricter lower bound, which is proven as follows.} 

\new{The scaling $\frac{||\boldsymbol{\mathcal{V}}||}{||\boldsymbol{U}\boldsymbol{V}^\top\dot{\boldsymbol{q}}||}$ cannot exceed~$\sigma_{\mathrm{max}}$. Assuming a consistent unit system, as discussed in~\secref{sc:fk},~$\boldsymbol{U},\boldsymbol{V}^\top$ act as pure rotations (no scaling) and we may use~$||\boldsymbol{\mathcal{V}}||,||\dot{\boldsymbol{q}}||$ to refer to their numerical values\footnote{\new{Noting that the units of~$\boldsymbol{v}$ and~$\boldsymbol{\omega}$ within~$\boldsymbol{\mathcal{V}}$ are themselves different, the norm~$\boldsymbol{\mathcal{V}}$ is defined only because we use~$\reals{6}$ as a local parameterization of~$\sethree$ (it is not a norm on~$\sethree$).}}, writing the scaling limit as~${\frac{||\boldsymbol{\mathcal{V}}||}{||\dot{\boldsymbol{q}}||} \leq \sigma_\mathrm{max}}$. Therefore, a lower bound on~$\sigma_{\mathrm{max}}$ can be derived by identifying a motion that is feasible at any arbitrary configuration, and calculating~${\sigma_{\mathrm{max}} \geq \frac{||\boldsymbol{\mathcal{V}}||}{||\dot{\boldsymbol{q}}||} = \frac{\sqrt{||\boldsymbol{v}||^2 + ||\boldsymbol{\omega}||^2}}{||\dot{\boldsymbol{q}}||}}$ for such a motion (for a spatial manipulator with~${m=6}$). If the task space is defined by translational DoFs alone, then~${\sigma_{\mathrm{max}} \geq \frac{||\boldsymbol{v}||}{||\dot{\boldsymbol{q}}||}}$ . }

For example, for a spatial manipulator \new{where the most distal joint is revolute}, \new{let~$l$ be the perpendicular distance between the end-effector point and the last joint axis (along a vector~$\boldsymbol{l}$, expressed in the aforementioned consistent unit system).} If the last joint alone is actuated with an angular velocity~$\dot{q}_n$ about the \new{unit }joint axis~${\hat{\boldsymbol{\omega}}_n}$, then the end-effector has~\new{${\boldsymbol{v} = \dot{q}_n (\hat{\boldsymbol{\omega}}_n \times \boldsymbol{l})}$, and~${\boldsymbol{\omega} = \dot{q}_n \hat{\boldsymbol{\omega}}_n}$}. Moreover, such a rotation is always feasible at \emph{any} configuration. Therefore,~${\sigma_{\mathrm{max}} \geq \new{\frac{||\boldsymbol{\mathcal{V}}||}{||\dot{\boldsymbol{q}}||}  =} \sqrt{1 + l^2}}$.

If the last joint is prismatic, then translation along its axis is feasible always, resulting in a unit scaling from~$||\dot{\boldsymbol{q}}||$ to~$||\boldsymbol{\mathcal{V}}||$, therefore~${\sigma_{\mathrm{max}} \geq \new{\frac{||\boldsymbol{\mathcal{V}}||}{||\dot{\boldsymbol{q}}||} =} 1}$ \new{(assuming joint velocity and end-effector velocity have the same units)}. In either case, for the combined linear and angular velocity Jacobian matrix of a serial manipulator,~${\sigma_{\mathrm{max}}}$ cannot fall below~$1$. 

\section{Modifications}
\label{sc:mods}
The algorithm presented in \secref{sc:mainalg} may be modified to address various considerations, as follows. 
\subsection{Time taken in singular regions}
If motion in the singular directions is slower than desired, or appears to be slowing down farther away from the singularity than necessary, either~$\gamma$ can be decreased or the profile of~$\Phi_i(\frac{\sigma_i}{\gamma \sigma_{\mathrm{max}}})$, and therefore~$S_i(\frac{\sigma_i}{\gamma \sigma_{\mathrm{max}}})$, can be varied. Defining~${\xi = \frac{\sigma_i}{\gamma \sigma_{\mathrm{max}}}}$; in the standard version of J-PARSE, the function used for entries corresponding to singular directions in the diagonal matrix~$\boldsymbol{S}$ is (from~\eqref{eq:sij}): 
\begin{align}
    S_i(\xi) = \Phi_i(\xi) = \xi.
\end{align}
The above may be replaced by any~$S_i(\xi)$ satisfying the following conditions: \textbf{(1)}~${S_i(\xi = 0) = 0}$, so that~${\dot{\boldsymbol{q}}=\boldsymbol{0}}$ at singularities; \textbf{(2)}~${S_i(\xi = 1) = 1}$, so that there is continuity at transitions into the near-singular region; \textbf{(3)}~$S_i^\prime(\xi) < \infty, \quad \forall \xi \in [0, 1]$, so that derivatives are bounded for stability; and \textbf{(4)}~$0 < S_i^\prime(\xi), \quad \forall \xi \in [0, 1]$, so that~$0 < S_i(\xi) < 1 \forall \xi \in (0,1)$, i.e., the maximum value of~$S_i$ occurs at the transition into the near-singular region, not within it. This ensures that the algorithm for computing~$\boldsymbol{S}$ does not result in positive acceleration along singular directions (barring a sharp drop in~$\sigma_{\mathrm{max}}$ within the region, which would exaggerate~$\Sigma_{s_i}^+$ and dominate the product~$\Sigma_{s_i}^+ S_i$ faster than is counteracted by the fall in~$S_i$).

An example of a simple function satisfying the above conditions is
\begin{align}
    S_i(\xi) = \frac{\xi(1+ a)}{1 + a \xi}, \quad a \in \reals{}_+
\end{align}
where~${a = 0}$ for the standard version. The derivative is bounded w.r.t.~$\xi$ (and therefore w.r.t.~$\boldsymbol{q}$, following from the argument in \secref{sc:contproof}):
\begin{align}
    S_i^\prime(\xi) = \frac{1 + a}{(1+a \xi)^2} 
\end{align}
As~$a, \xi \geq 0$, it is clear that~$S_i^\prime(\xi)$ is positive and bounded on~$[0,1]$. For~$a\ne0$, more of the deceleration occurs closer to singularity, i.e., as~$\xi$ decreases. Increasing~$a$ exaggerates this effect. Extremely large values of~$a$ should be avoided, as they would result in large derivatives close to~$\xi = 0$. In fact, as~${a \rightarrow \infty}$, the entirety of the deceleration is demanded at~$\xi = 0$, \new{violating Condition 3 above}. That is,~$S_i^\prime(\xi = 1) \rightarrow 0$ and ~$S_i^\prime(\xi = 0) \rightarrow \infty$. In practice, it was observed that~$a$ can be varied without concern among smaller values. The behavior is thus tunable in terms of the extent to which it approaches singularities before significantly slowing down, and at the same time, is not overly sensitive to the tuning parameter in terms of stability. 

\subsection{Task-space coupling}
\label{sc:taskspacecoup}
The columns of~$\boldsymbol{U}$ do not, in general, correspond to clearly-separated position or orientation \gls{DoF}s in the task space. This is a common feature of \gls{SVD}-based methods, including DLS, adaptive DLS, and EDLS. In fact, in general, it is not always possible to classify a singularity in~$\boldsymbol{J}$ as arising from either~$\boldsymbol{J_v}$ or~$\boldsymbol{J_\omega}$ alone. Nevertheless, situations may arise where it is of interest to separately control the behavior in dimensions of the task space, rather than along the principal axes of the manipulability $m$-ellipsoid.

\textit{Linear scaling in task space}: Rather than \ins{varying the function~$\Phi_i(\sigma_i)$}, a variable gain matrix may be introduced in the task space, modifying \eqref{eqn:jparse_full} to~${\dot{\boldsymbol{q}} = (\boldsymbol{U}\boldsymbol{U}_p^{+} + \tilde{\boldsymbol{U}} \boldsymbol{\Phi} \tilde{\boldsymbol{U}}^\top \boldsymbol{K}_s) \boldsymbol{\mathcal{V}}}$. As this disrupts the diagonal form of~$\boldsymbol{S}$ derived in \eqref{eqn:jparse_simples}, further analysis is left for future work. For continuity, it is important that~$\boldsymbol{K}_s$ equals the identity matrix at the transition point.

\textit{Task priority for linear/angular velocity}: In scenarios where the separation of position and orientation is important, a task-priority framework may be considered, with J-PARSE being implemented on a subspace of~$\sethree$. For example, if accuracy in~$\sothree$ is prioritized over that in~$\reals{3}$, and a singularity is encountered in~$\reals{3}$, then~${\dot{\boldsymbol{q}} = \boldsymbol{J}_\omega^+ \boldsymbol{\omega} + (\boldsymbol{I} - \boldsymbol{J}_{\omega}^+\boldsymbol{J}_{\omega}) \boldsymbol{J}_{v}{}_{ \mathrm{parse}}^+ \boldsymbol{v}}$. Such an approach cannot be explored without more thorough analysis, hence its study is left for future work. 

\subsection{Redundancy and branch switching}
\label{sc:branches}
It is common in redundant manipulators to obtain a~$\dot{\boldsymbol{q}}$ corresponding to the secondary objective and project it into the nullspace of~$\boldsymbol{J}$ using the projection matrix~${(\boldsymbol{I} - \boldsymbol{J}^+ \boldsymbol{J})}$. Singularity-handling methods recognize that the task-space effect of this term no longer vanishes if~$\boldsymbol{J}^+$ is replaced with a modified inverse~\cite{deo1995}, which is likewise true for J-PARSE. For examples involving redundant manipulators in the present work, we use a projection into the nullspace of~$\boldsymbol{J_s}$ instead, for example:
\begin{align}
    \boldsymbol{J}^+_{\mathrm{parse}} = \boldsymbol{V} \boldsymbol{\Sigma}_s^+ \boldsymbol{U}^\top \boldsymbol{\mathcal{V}} +  (\boldsymbol{I} - \boldsymbol{J}_s^+ 
    \boldsymbol{J}_s) \boldsymbol{\nabla}H(\boldsymbol{q}),
\end{align}
where~$H(\boldsymbol{q})$ is a potential field in the joint-space, driving attraction to a desirable nominal \new{configuration}. The study of other variations of this term, and stability analysis with consideration of redundancy, are left for future work. The inclusion of this term has the additional advantage of biasing the motion towards a chosen branch when the robot is exiting a singular configuration. Without such a term, approaching the singular configuration in practical settings, even safely, would be a poor decision, since the manipulator may \ins{move from the singularity into }any of the branches merging at that configuration -- some of which would make collision and joint-limit avoidance more inconvenient than others. 

\section{Results}
\label{sc:results}

The experiments reported in this work are distributed over a variety of architectures, aiming to cover characterization, ablation, benchmarking, and demonstration of applications. Since J-PARSE is designed for online control with a human in the loop, the last section shows practical examples with human operators. However, a uniform set of programmed inputs is used for the sections examining effects of the terms and the parameters, and for comparisons to other methods. 

\textbf{A. Boundary singularities of planar~$2$-R manipulator}: First, J-PARSE is characterized in \secref{sc:2r} using a kinematic simulation of the planar~$2$-R manipulator, as it is the simplest possible serial manipulator capable of experiencing \new{IK} singularities. This simplicity is useful in observing the effects of tuning parameters for J-PARSE, DLS, \new{ADLS} and EDLS. Under proportional control with uniform gain in the task space, and with uniform starting configurations across the algorithms, the end-effector is driven (a) into a singular goal configuration, and (b) out of it. This provides intuition for the tuning of all algorithms, and the relative difficulty of tuning ADLS. 

\textbf{B. Boundary singularities of spatial manipulator}: Next, to demonstrate handling of singular and near-singular configurations beyond the simple~$2$-\gls{DoF} example, discrete goals in~$\sethree$ are set for the end-effector of a UFactory X-Arm7, a redundant spatial manipulator, creating a series of keypoints, some of them within and some outside of the workspace. Simulating the manipulator dynamics in \gls{ROS} Gazebo~\cite{koeniggazebo}, the reaching behavior under J-PARSE is shown in contrast to both, variously-tuned iterations of DLS, and ablations of J-PARSE itself. The DLS comparisons serve to reiterate that damping causes either non-zero steady state errors away from singularities, or instabilities near them. The ablation studies demonstrate the role of each term in the J-PARSE algorithm, by showing the effects on behavior if it is left out. The same series of keypoints is then reached by a physical X-Arm7 controlled with J-PARSE. Selected results from this study are reported in~\secref{sc:goalsreach}.

\textbf{C. Internal singularities of spatial manipulators}: Spatial manipulators also exhibit singularities ``internal'' to the reachable workspace. Such singularities are not necessarily approached while reaching towards unreachable goal poses, but may be even crossed while traversing an end-effector path seemingly contained within the reachable volume. To investigate behavior near such configurations, in~\secref{sc:intsing}, two coincident-wrist spatial manipulators are \new{used}: one redundant (Kinova Gen3 manipulator) and one non-redundant (PUMA 560 manipulator). \new{The former was teleoperated, while the latter was simulated in Gazebo with a} pre-programmed straight-line path in the task space. In a practical closed-loop teleoperation setting, a human operator would give task-space inputs based on the current pose of the end-effector, accounting for any deceleration or deviations near singularities. To approximate this with an open-loop test setup, the commanded goal pose moves slowly back and forth along the path. 

\textbf{D. Practical examples of online control}: Finally, in~\secref{sc:demos}, J-PARSE is used to demonstrate task-space teleoperation of the XArm7 manipulator for a pick-and-place task. The same task demonstrated with the XArm is repeated to collect demonstrations for an imitation learning algorithm, showing that the task may then be performed following task-space twists predicted by a diffusion policy. As another example, the Gen3 manipulator, which has a camera at the end-effector, is used for a visual servoing task.

\subsection{Boundary singularities of planar~\texorpdfstring{$2$}{2}-R manipulator}
\label{sc:2r}
\begin{figure*}[htbp]
     \centering
     \def\svgwidth{0.8\textwidth} 
        {\sffamily
        \tiny
        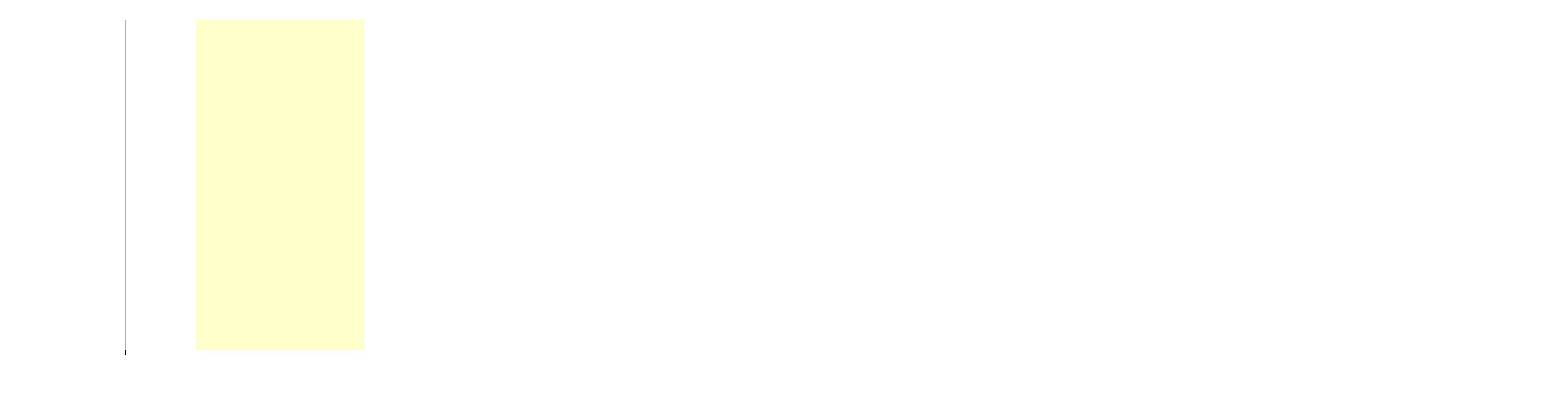
        }
     \caption{\ins{Simulation results for the joint velocity~$\dot{q}_2$ of the planar~$2$-R manipulator moving from a non-singular to singular configuration, controlled by the algorithms described in Table~\ref{tab:2r} (a) DLS and ADLS, (b) J-PARSE and EDLS, (c) J-PARSE and ADLS tuned for similar behavior. All controllers used~$k = 0.1,~\Delta t = 0.01$~s.}}
    \label{fig:2r_comp}
    \vspace{-1.5em}
\end{figure*}

A kinematic simulation ($\Delta t = 0.01$) of planar~$2$-R manipulator with unit link lengths is placed under proportional control in the task space (under a gain of~$0.1$) from a non-singular configuration ($q_1 = -\frac{\pi}{4}, q_2 = \frac{\pi}{4}$) towards an unreachable pose ($x = y = 1.1\sqrt{2}$). The algorithms in Table~\ref{tab:2r} are compared in \figref{fig:2r_comp}. In each case,~${\dot{\boldsymbol{q}} = \boldsymbol{V} \boldsymbol{Y} \boldsymbol{U}^\top \new{\boldsymbol{e}}}$, where the entries of~$\boldsymbol{Y}$ characterize the algorithm being used. 

\new{In \figref{fig:2r_comp}a and \figref{fig:2r_comp}b, parameters are varied over a range to illustrate the influence of each relevant parameter. For DLS, the values in Table~\ref{tab:2r} correspond to~${\lambda = \sqrt{0.05}, \sqrt{0.03}, \sqrt{0.01}}$). For ADLS, the central value of~$\lambda$ from the DLS examples is chosen, varying~$w_0$ alone to show its effect. For EDLS,~${\sigma^+, \beta}$ are taken from~\cite{carmichaelFrameworkSingularityrobustManipulator2017} and~$\sigma^-$ is set to zero so as not to explicitly avoid the singularity. For J-PARSE,~$\gamma$ is varied at intervals which allow sufficiently distinct behavior for easy visual interpretation. }
\vspace{-0.5em}
\begin{table}[htbp]
\centering
\caption{Algorithms and parameter settings compared on the planar~$2$-R manipulator. ADLS uses the same~$Y_i$ as DLS, with~$\lambda_0 = 0.17$, but with adaptive damping~($w$ being the manipulability measure).}
\label{tab:2r}
\begin{tabular}{|c|c|c|}
\hline
\textbf{Algorithm} & \textbf{Inverse singular value $Y_i$} & \textbf{Parameters} \\ \hline
\multirow{3}{*}{DLS~\cite{nakamuraInverseKinematicSolutions1986, wamplerManipulatorInverseKinematic1986}} & \multirow{3}{*}{$\frac{\sigma_i}{\sigma_i^2 + \lambda^2}$, $\lambda$ constant} & $\lambda = 0.22$ \\ \cline{3-3} 
                   &                              & $\lambda = 0.17$ \\ \cline{3-3} 
                   &                              & $\lambda = 0.10$ \\ \hline
\multirow{3}{*}{ADLS~\cite{nakamuraInverseKinematicSolutions1986}} & \multirow{3}{*}{$\lambda = \begin{cases} \lambda_0 \left(1 - \frac{w}{w_0}\right) \textrm{if} w < w_0 \\
0, \textrm{otherwise}
\end{cases}$} & $w_0 = 0.50$ \\ \cline{3-3} 
                   &                              & $w_0 = 0.25$ \\ \cline{3-3} 
                   &                              & $w_0 = 0.10$ \\ \hline
\multirow{3}{*}{EDLS~\cite{edls2017}} & \multirow{3}{*}{$\frac{1 - \beta^{\left(\frac{\sigma_i - \sigma^{-}}{\sigma^{+} - \sigma^{-}}\right)}}{\sigma_i}$} & $\sigma^{-} = 0.00$, \\ 
                   &                              & $\sigma^{+} = 0.30$, \\ 
                   &                              & $\beta = 0.02$ \\ \hline
\multirow{3}{*}{J-PARSE} & \multirow{3}{*}{$\begin{cases}
\frac{\sigma_i}{(\gamma\sigma_{\mathrm{max}})^2} \textrm{ if} \sigma_i < \gamma \sigma_{\mathrm{max}}, \\
\frac{1}{\sigma_i} \textrm{ otherwise}
\end{cases}
$} & $\gamma = 0.10$ \\ \cline{3-3} 
                   &                              & $\gamma = 0.06$ \\ \cline{3-3} 
                   &                              & $\gamma = 0.03$ \\ \hline
\end{tabular}
\vspace{-0.5em}
\end{table}
\begin{figure*}[htbp]
     \centering
     \def\svgwidth{0.8\textwidth} 
        {\sffamily
        \tiny
        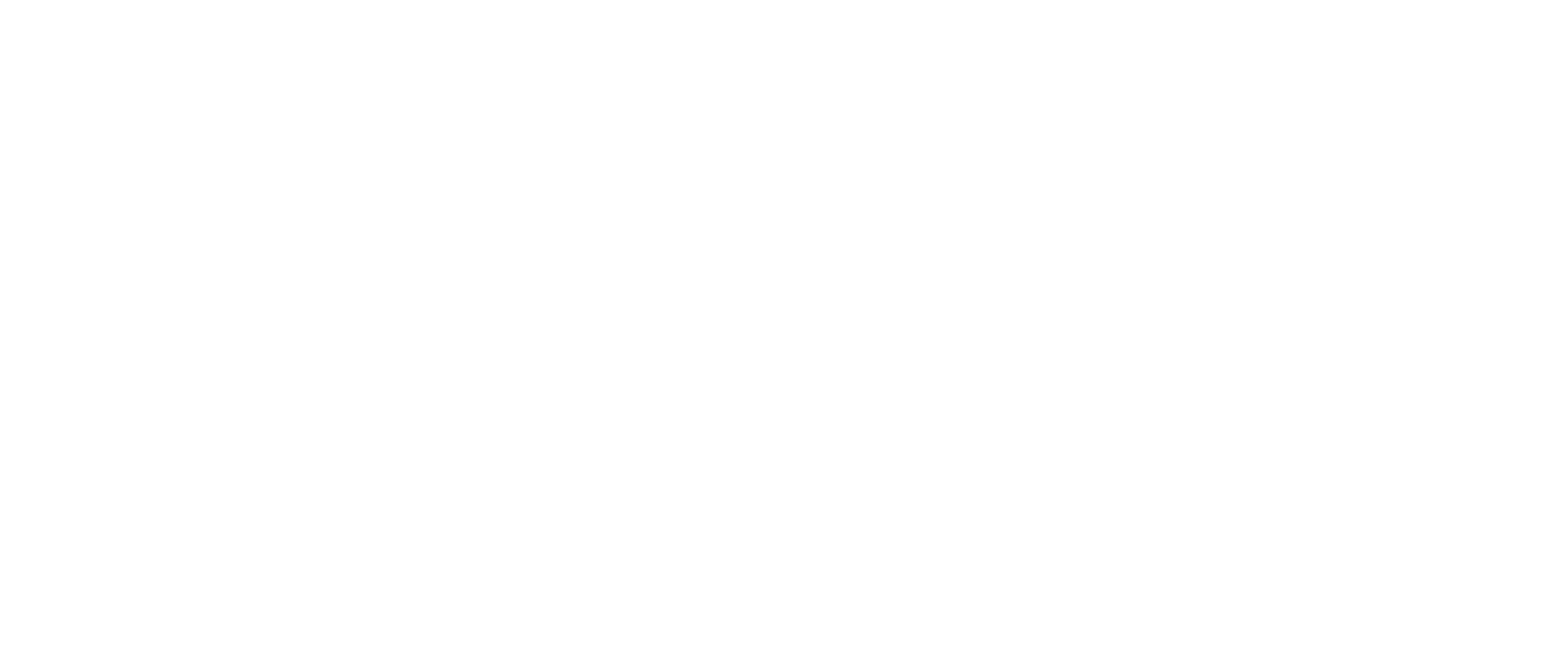
        }
     \caption{\ins{Simulation results for the planar~$2$-R manipulator moving from a (a) non-singular to singular configuration identical to~\figref{fig:2r_comp} (parameters of ADLS examples are tuned to visually match the behavior of J-PARSE examples); (b) singular to non-singular configuration (all three previously-tuned ADLS parameters result in unbounded joint speeds). At the singular configuration,~$q_2 =0$. All controllers used~$k = 0.1,~\Delta t = 0.01$~s.}}
    \label{fig:2r_in_out}
\end{figure*}

EDLS fails to achieve stable behavior with~${\sigma^- = 0}$, as ${\lim_{\sigma_i \rightarrow 0}S_i = -\frac{\ln\beta}{\sigma^+} \ne 0}$. With DLS, as~$\lambda$ is decreased, the response becomes less sluggish; ADLS, with adaptive~$\lambda$, performs even better. The joint velocity profiles with ADLS are differentiable everywhere, while those with J-PARSE are not, however, the behavior is stable, \ins{as discussed in \secref{sc:vdot}. }As expected, decreasing~$\gamma$ in J-PARSE shrinks the region of modified behavior, and results in higher peak joint speeds at the transition into the singular region. This is similar to the effect of decreasing~$w_0$ in ADLS; in fact, with careful tuning, very similar behaviors for a given trajectory can be achieved with J-PARSE and ADLS, \ins{as shown in~\figref{fig:2r_comp}c}. \new{\figref{fig:2r_comp}c presents a comparison where ADLS is tuned to achieve visually similar joint trajectories to each of the three examples of J-PARSE for the selected target pose. Selected results are given in Table~\ref{tab:accjerks}, showing comparable final errors and} peak joint accelerations between the two algorithms, and higher jerks for J-PARSE, as expected. However, the parameter selection process for using ADLS requires \ins{variation} of two parameters~$\lambda_0$ and~$w_0$. \figref{fig:2r_in_out} illustrates the practical implications of this difficulty in tuning\new{, re-using the two pairs of tuning settings from \figref{fig:2r_comp}c that resulted in lower peak velocities}. Here, the goal pose is made reachable (set to~$(0.5\sqrt{2}, 0.5\sqrt{2})$) but the starting configuration is set to ($q_1 = \frac{\pi}{4}, q_2 = 10^{-10}$). It is observed that the same ADLS parameters that yield similar behavior to J-PARSE, for the initial goal in~\figref{fig:2r_in_out}a, now result in joint speeds on the order of \new{thousands of}~rad/s\new{, and even higher accelerations and jerks (Table~\ref{tab:accjerks_exit}),} whereas J-PARSE is able to guide the manipulator stably out of the near-singular starting configuration and to the goal~(\figref{fig:2r_in_out}b). \new{These demonstrations suggest, consistent with~\cite{devito2017}, that \textbf{although ADLS can be tuned to yield similar behavior as J-PARSE for a given situation, ADLS requires per-situation retuning whereas J-PARSE does not}.}
\begin{table}
    \centering
    \caption{\new{Minimum ICN, peak values of velocity, acceleration and jerk for joint 1 and joint 2 when going towards a pose~$200$~mm outside the workspace (\figref{fig:2r_comp}). Final error from target was~$200.001$~mm in all cases.}}
    \label{tab:accjerks}
    \begin{tabular}{|l|l|r|r|r|}
    \hline
        Algorithm & ICN$_{\mathrm{min}}$ &$\dot{q}_1$~and~$\dot{q}_2$ & $\ddot{q}_1$~and~$\ddot{q}_2$ & $\dddot{q}_1$~and~$\dddot{q}_2$ \\
        (parameters) & & (rad/s) & (rad/s$^2$) & (rad/s$^3$)\\
        \hline
        ADLS & $0.009$ & $0.057$ & ${0.005}$ & $0.001$ \\
        ($0.26,1.00$) & & $0.070$ & $0.007$ & $0.002$ \\
        \hline
        ADLS & $0.003$ & $0.070$ & $0.019$ & $0.019$ \\
        ($0.22,0.45$) & & $0.106$ & $0.034$ & $0.038$ \\
        \hline
        ADLS & $0.001$ & $0.143$ & $0.408$ & $6.386$ \\
        ($0.17,0.15$) & & $0.258$ & $0.812$ & $12.76$ \\
        \hline
        J-PARSE & $0.006$ & $0.066$ & $0.017$ & $0.756$ \\
        ($0.10$) & & $0.094$ & $0.029$ & $1.495$ \\
        \hline
        J-PARSE & $0.001$ & $0.083$ & $0.041$ & $2.413$ \\
        ($0.06$) & & $0.135$ & $0.078$ & $4.805$ \\
        \hline
        J-PARSE & $0.000$ & $0.139$ & $0.227$ & $15.48$ \\
        ($0.03$) & & $0.249$ & $0.449$ & $30.93$ \\
        \hline
    \end{tabular}
    \vspace{-1.6em}
\end{table}

\begin{table}
    \centering
    \vspace{-0.5em}
    \caption{\new{Final error, peak values of velocity, acceleration and jerk for joint 1 and joint 2 when starting from a singular configuration (\figref{fig:2r_in_out}). Starting ICN was~$0.000$ for all cases.}}
    \label{tab:accjerks_exit}
    \begin{tabular}{|l|l|r|r|r|}
    \hline
        Algorithm & Error & $\dot{q}_1$~and~$\dot{q}_2$ & $\ddot{q}_1$~and~$\ddot{q}_2$ & $\dddot{q}_1$~and~$\dddot{q}_2$ \\
        (parameters) & (mm) & & (rad/s$^2$) & (rad/s$^3$) \\
        \hline
        ADLS & $0.316$ & $4.637\times10^3$ & $2.318\times10^5$ & $2.318\times10^7$ \\
        ($0.26,1.00$) & & $9.273\times10^3$ & $4.637\times10^5$ & $4.637\times10^7$ \\
        \hline
        ADLS & $1.060$ & $2.308\times10^3$ & $1.154\times10^5$ & $1.154\times10^7$ \\
        ($0.22,0.45$) & & $4.615\times10^3$ & $2.307\times10^5$ & $2.308\times10^7$ \\
        \hline
        ADLS & $2.386$ & $4.566\times10^3$ & $4.566\times10^5$ & $3.424\times10^7$ \\
        ($0.17,0.15$) & & $9.131\times10^3$ & $9.131\times10^5$ & $6.848\times10^7$ \\
        \hline
        J-PARSE & $5.161$ & $0.193$ & $0.162$ & $11.71$ \\
        ($0.10$) & & $0.385$ & $0.323$ & $23.13$ \\
        \hline
        J-PARSE & $0.882$ & $0.326$ & $0.694$ & $51.91$ \\
        ($0.06$) & & $0.651$ & $1.376$ & $103.4$ \\
        \hline
        J-PARSE & $0.426$ & $0.637$ & $4.959$ & $355.1$ \\
        ($0.03$) & & $1.274$ & $9.902$ & $709.8$ \\
        \hline
    \end{tabular}
    \vspace{-0.5em}
\end{table}

\subsection{Boundary singularities of spatial manipulator}
\label{sc:goalsreach}
\begin{figure}[h]
\centering
\begin{minipage}[c]{0.3\columnwidth}
    \raggedleft
    \begin{tabular}{|c|c|c|c|}
    \hline
    \textbf{L} & \textbf{$x$} & \textbf{$y$}  & \textbf{$z$} \\ \hline
    A & 1.00 & 0.00 & 0.50 \\ 
    B & 0.50 & 0.00 & 0.50 \\ 
    \rowcolor{gray!15}
    C & 0.00 & 0.00 & 0.50 \\ 
    D & 0.50 & 0.00 & 0.50 \\ \hline
    \textbf{P} & \textbf{$x$} & \textbf{$y$}  & \textbf{$z$} \\ \hline
    A & 0.55 & 0.00 & 0.30 \\ 
    \rowcolor{gray!15}
    B & 0.40 & 0.80 & 0.30 \\ 
    C & 0.25 & 0.00 & 0.30 \\ 
    \rowcolor{gray!15}
    D & 0.40 & -0.80 & 0.30 \\ \hline
    \end{tabular}
    \label{tab:abcd}
\end{minipage}
\hspace{4em}
\begin{minipage}[c]{0.5\columnwidth}
    \raggedright
    \includegraphics[width=\textwidth, trim=0 50 0 0, clip]{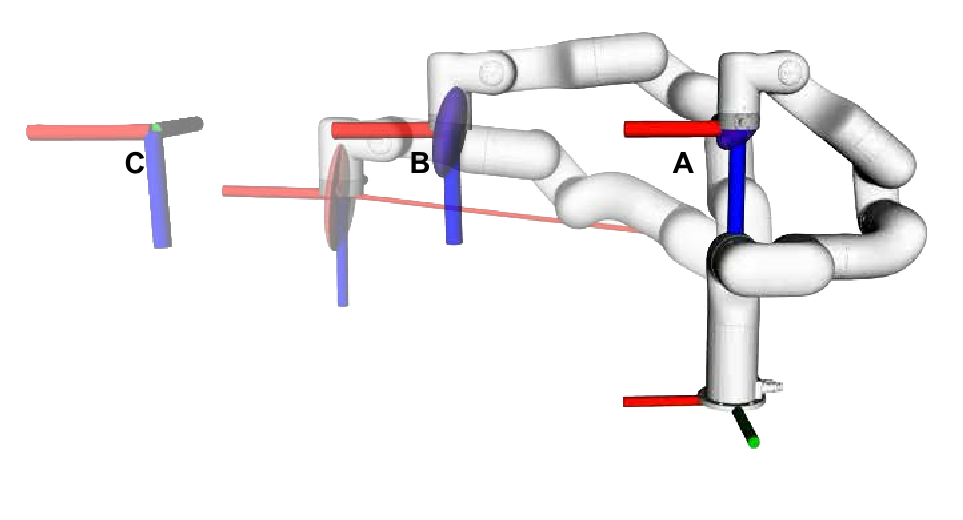}
    \raggedright
    \includegraphics[width=\textwidth, trim=0 50 0 0, clip]{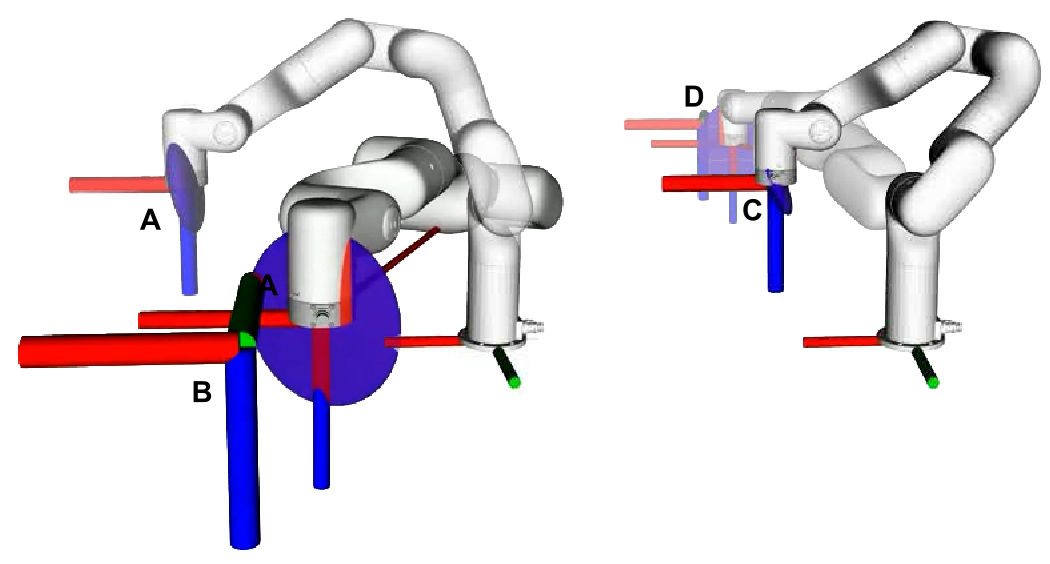}
\end{minipage}
\caption{Two sets of keypoints were used for the goal reaching experiments with X-Arm (L: line, P: plane). All target poses had the same orientation. Unreachable poses are shaded.}
\label{fig:keypoints}
\vspace{-0.5em}
\end{figure}

The~$7$-\gls{DoF} UFactory X-Arm was driven under proportional control to keypoints listed in~\figref{fig:keypoints}, one set lying on a straight line and the other spread over a horizontal plane. Proportional gains were selected as~$k_{\mathrm{pos}} = 10, k_{\mathrm{ori}} = 10$ at a controller frequency of~$50$~Hz, thus \textbf{satisfying the stability condition}~$k\Delta t = 0.2 <2$ from \secref{sc:discproof}. For the real robot, unit gains were used \ins{($k \Delta t = 0.02 < 2$)}. In a practical teleoperation setting, the magnitude of the instantaneous commanded twist~$\boldsymbol{\mathcal{V}}$ would be limited by safety considerations, and not proportional to the (unbounded) error from the goal pose. To reflect this,~$||\boldsymbol{\mathcal{V}}||$ was capped at~$1$ in the simulation experiments and at~$0.1$ in the real robot experiments. \new{Bounding the task-space speeds within reasonable limits is required for safety, even far away from singularities (for all algorithms). For J-PARSE, it also ensures that the local treatment (Euclidean approximation of~$\sothree$ in \secref{sc:contproof}, first-order approximation of system dynamics in \ref{sc:discproof}, finite joint speeds in \secref{sc:gamma}) remains valid.} The nullspace objective was defined as~${\boldsymbol{\nabla} H(\boldsymbol{q}) = \new{\boldsymbol{\eta}} = -k_n\new{(\boldsymbol{q} - \boldsymbol{q}_{\mathrm{nom}})}}$, \new{using the nominal configuration~$\boldsymbol{q}_{\mathrm{nom}} = \boldsymbol{0}$.} Each entry of \new{$\boldsymbol{\eta}$} was capped at~$0.6$~rad/s ($k_n = 2$ for simulation,~$3$ for real, tuned by trial and error). \new{The nullspace objective ensures that the manipulator maintains an ``elbow-up'' configuration, both through redundancy resolution (relevant to all algorithms) and while exiting boundary singularities (for J-PARSE).} In each case, the robot was given a fixed time period to reach each pose. Since the conclusions from both sets of keypoints were similar, results for reaching the line keypoints are presented here, however the remaining results are available in the supplementary material.
The following algorithms were compared on the X-Arm:
\begin{enumerate}
    \item Jacobian pseudoinverse (using \eqref{eqn:Jpinv_svd}) \cite{whitneyMathematicsCoordinatedControl1972,chiaveriniKinematicallyRedundantManipulators2008}:
\begin{equation}
    \dot{\boldsymbol{q}} = \boldsymbol{J}^+ \boldsymbol{\mathcal{V}} + (\boldsymbol{I} - \boldsymbol{J}^+ \boldsymbol{J}) \boldsymbol{v}_{q}.
    \label{eqn:J_pinv_vel_full_control_with_nullspace}
\end{equation}
    \item Damped-Least-Squares pseudoinverse \cite{wamplerManipulatorInverseKinematic1986, chiaveriniKinematicallyRedundantManipulators2008}:
\begin{equation}
    \dot{\boldsymbol{q}} = \boldsymbol{J}^+_{\mathrm{DLS}} \boldsymbol{\mathcal{V}} + (\boldsymbol{I} - \boldsymbol{J}^+_{\mathrm{DLS}} \boldsymbol{J}) \boldsymbol{v}_{q}
    \label{eqn:J_dsl_vel_full_control_with_nullspace}
\end{equation}
As adaptive variants of \gls{DLS} involve setting user preferences for various parameters (such as the stopping distance away from singularity, the rate of adaptation of the~$\lambda$ and so on), \emph{it is most informative to benchmark directly against the original method}, \ins{having already investigated in~\secref{sc:2r} the difficulty in interpreting or predicting results from adaptive damping}. It is to be noted that neither \gls{DLS} nor its variants explicitly attempt to reach singular configurations. 
    \item JacobianProjection : To illustrate the role of each component in J-PARSE, incomplete versions of the algorithm are implemented, beginning with only the matrix $\boldsymbol{J}_p$ as developed in \secref{sc:projection_jacobian}: 
\begin{equation}
    \dot{\boldsymbol{q}} = \boldsymbol{J}_p^+ \boldsymbol{\mathcal{V}} + (\boldsymbol{I} - \boldsymbol{J}_p^+\boldsymbol{J}_p) \new{\boldsymbol{\eta}}
\end{equation}
    \item JacobianSafety: Next, the Safety Jacobian~$\boldsymbol{J}_s$ alone is used as a substitute for~$\boldsymbol{J}^+$: 
\begin{equation}
    \dot{\boldsymbol{q}} = \boldsymbol{J}_s^+ \boldsymbol{\mathcal{V}} + (\boldsymbol{I} - \boldsymbol{J}_s^+\boldsymbol{J}_s) \new{\boldsymbol{\eta}}
\end{equation}
    \item JacobianSafetyProjection: In this algorithm, only the term corresponding to the non-singular directions is used, the name being a shorthand for the sequence of matrices used in its construction: 
\begin{equation}
    \dot{\boldsymbol{q}} = \boldsymbol{J}_s^+ \boldsymbol{J}_p \boldsymbol{J}^+_p \boldsymbol{\mathcal{V}} + (\boldsymbol{I} - \boldsymbol{J}_s^+\boldsymbol{J}_s) \new{\boldsymbol{\eta}}
\end{equation}
\end{enumerate}
For J-PARSE and its ablations,~$\gamma = 0.1$ in simulation studies, \ins{so that the effect may be seen more clearly, and~$\gamma = 0.07$ on the real robot, to demonstrate a reasonable choice for that architecture (selected by observation as described in~\secref{sc:gamma})}. All poses are reported in the manipulator base frame, and the position and orientation errors are as defined in \secref{sc:forml}\footnote{where $\log_m(\cdot)$ is the matrix logarithm (inverse of the matrix exponential), and $\veemap((\cdot))$ is the `vee-map' and is the mathematical inverse of the skew-symmetric matrix `hat-map' operator~$\hatmap{(\cdot)}$}:
\begin{align}
    e_{x} &= ||\boldsymbol{p}_{\mathrm{des},xyz}-\boldsymbol{p}_{xyz} ||, \nonumber \\
    e_{{\theta}} &= || \hat{\boldsymbol{h}} \theta || = ||\veemap{\left(\log_{m}(\boldsymbol{R}_{\mathrm{des}} \boldsymbol{R}^\top)\right)}||. \nonumber
\end{align}

\begin{figure*}[htbp]
    \centering
    \begin{subfigure}[b]{0.49\textwidth}
        \centering
        \def\svgwidth{0.8\textwidth} 
        {\sffamily
        \tiny
        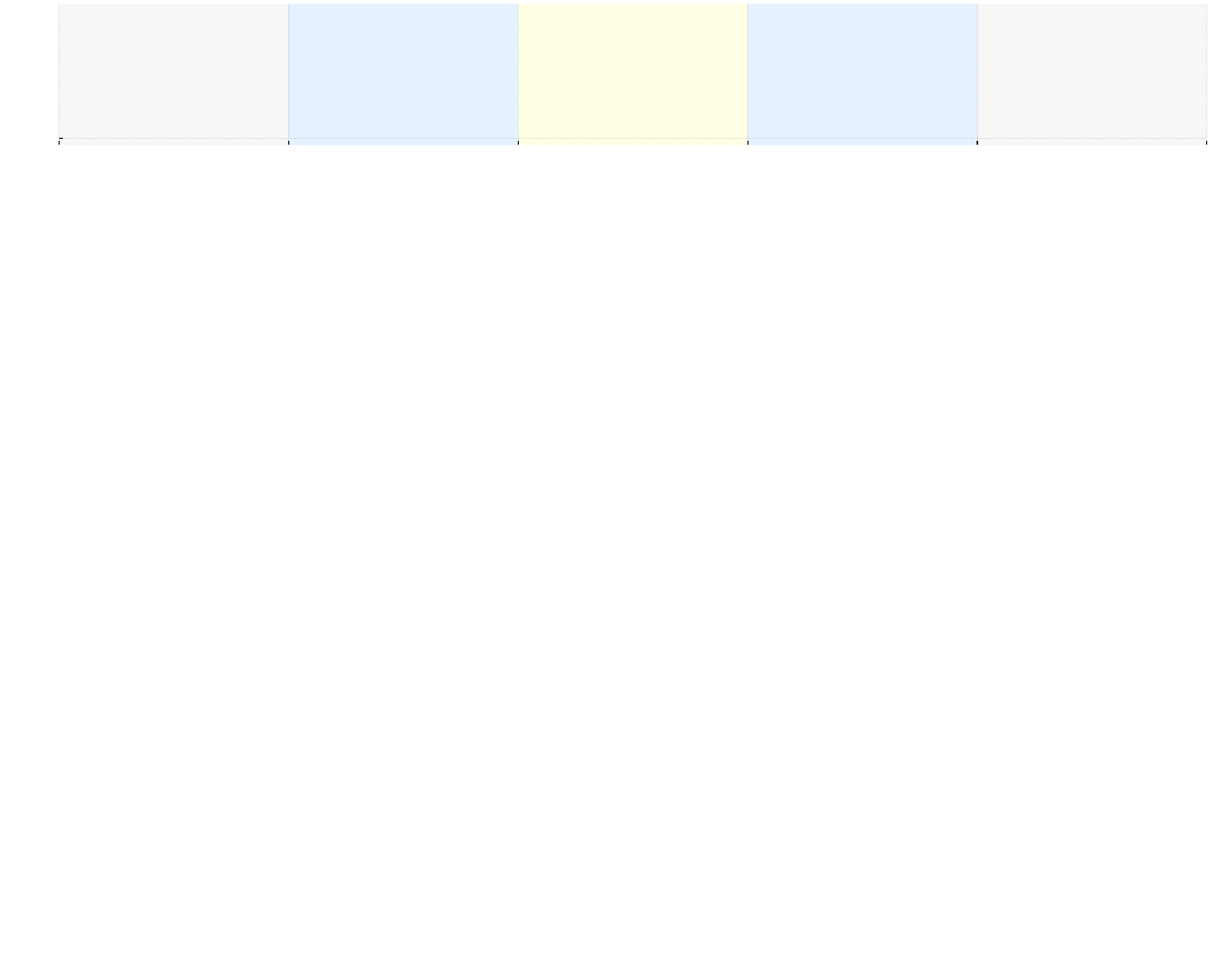
        }
         \caption{Regular pseudoinverse}
        \label{fig:xarm_jparse_v_pinv}
    \end{subfigure}
    \hfill
    \begin{subfigure}[b]{0.49\textwidth}
        \centering
        \def\svgwidth{0.8\textwidth} 
        {\sffamily
        \tiny
        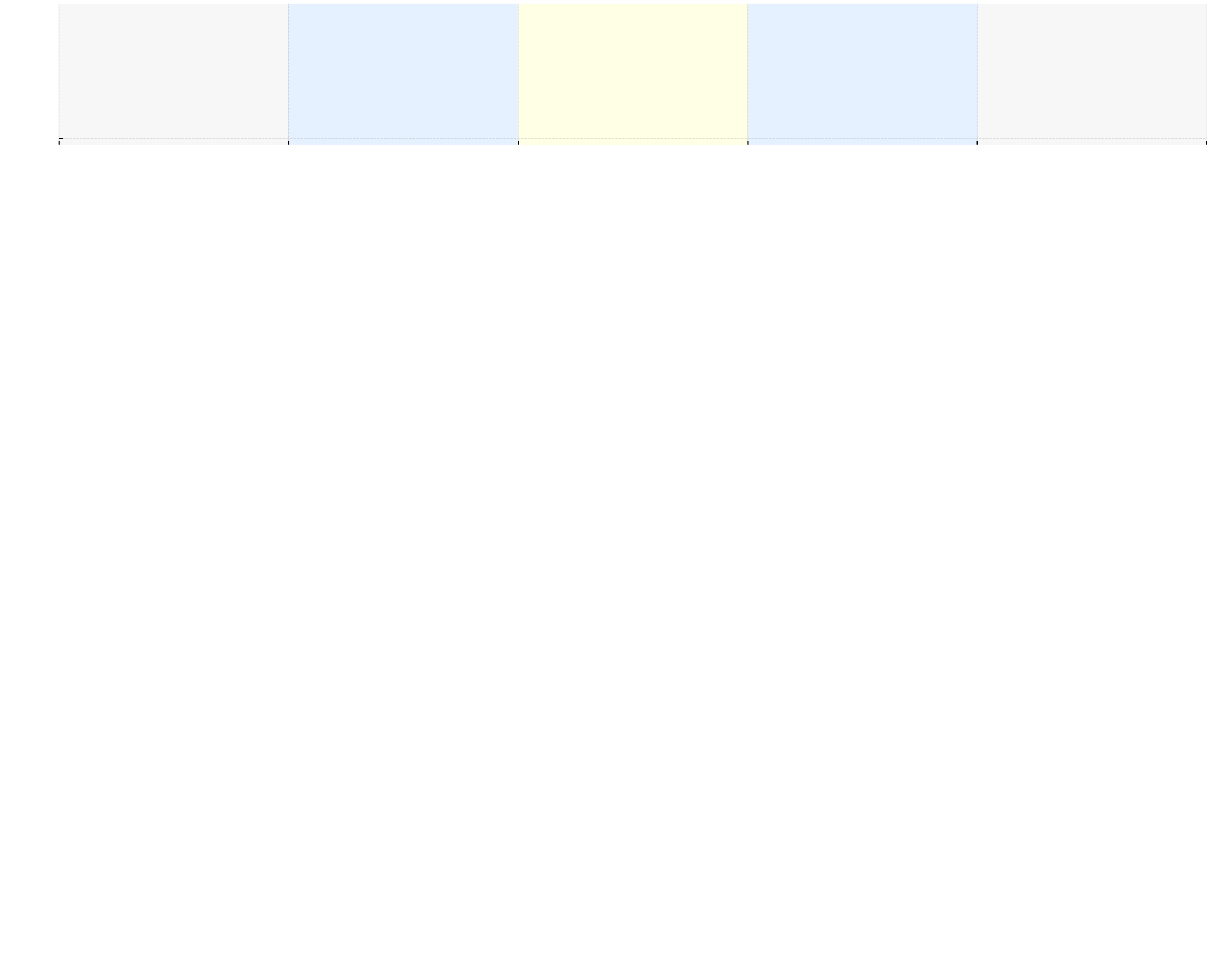
        }
        \caption{JacobianSafety (ablation of J-PARSE)}
        \label{fig:abl1}
    \end{subfigure}
    
    \vspace{0.5em}
    
    \begin{subfigure}[b]{0.49\textwidth}
        \centering
        \def\svgwidth{0.8\textwidth} 
        {\sffamily
        \tiny
        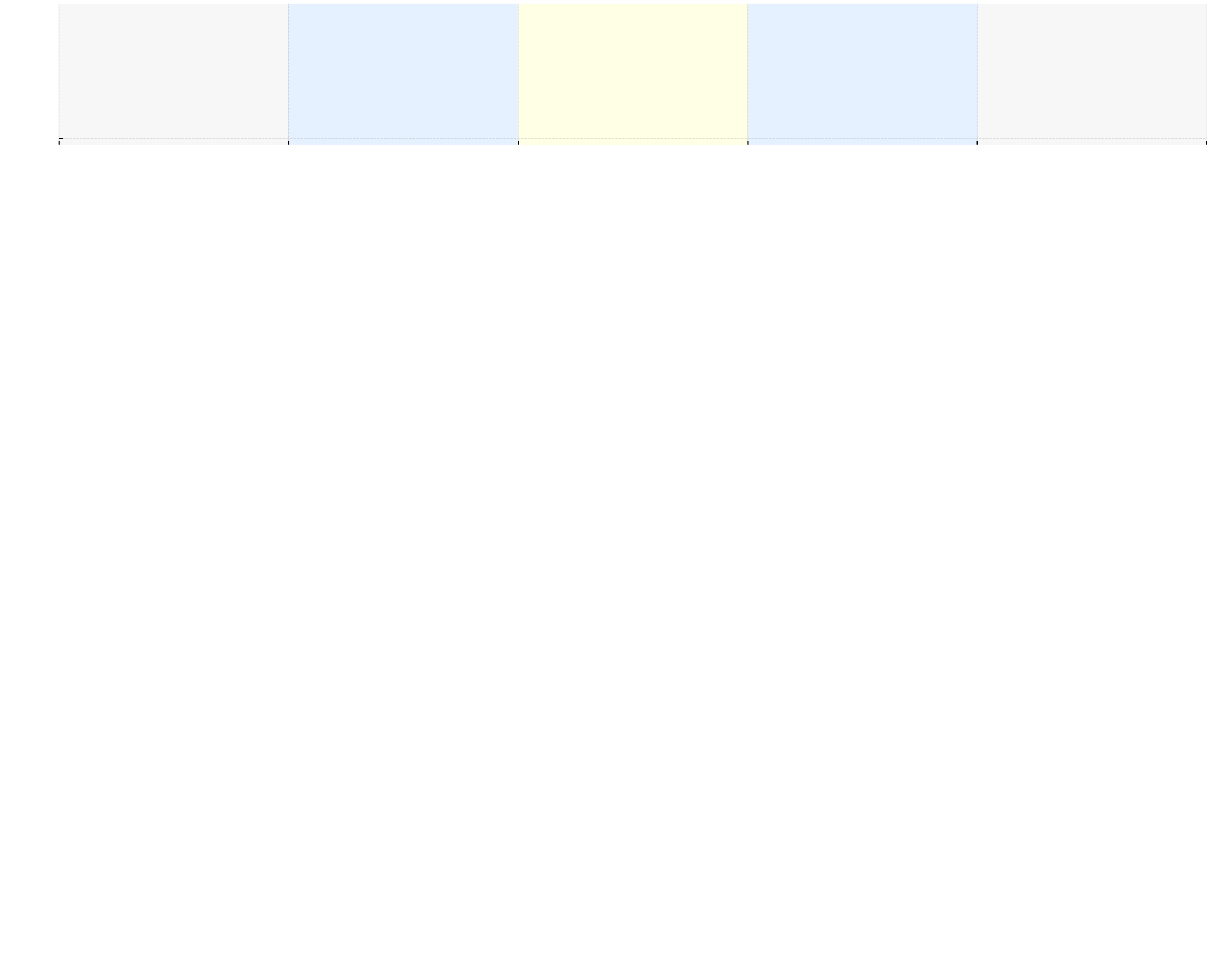
        }
         \caption{Various ablations of J-PARSE}
        \label{fig:abl2}
    \end{subfigure}
    \hfill
    \begin{subfigure}[b]{0.49\textwidth}
        \centering
        \def\svgwidth{0.8\textwidth} 
        {\sffamily
        \tiny
        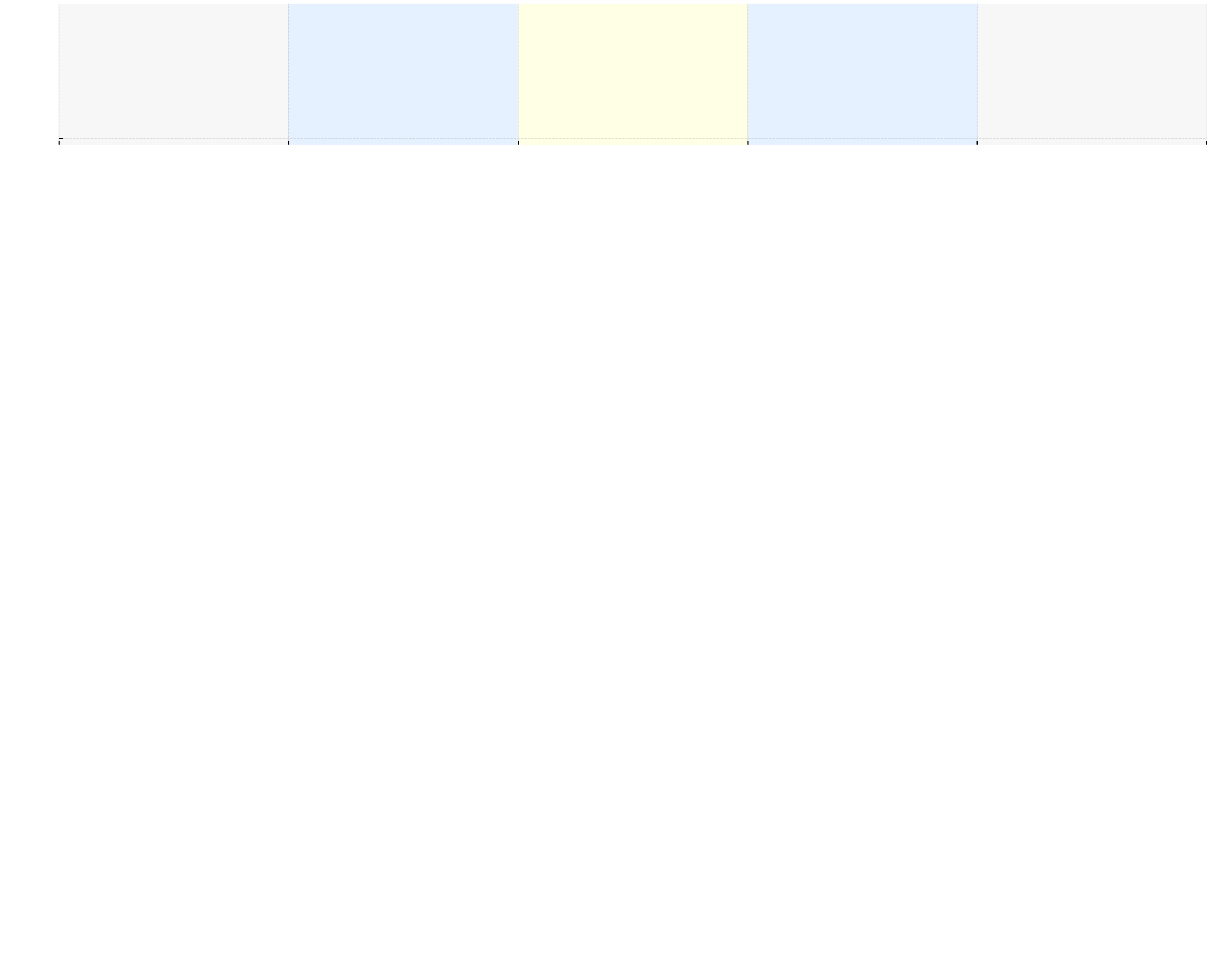
        }
        \caption{Damped Least Squares}
        \label{fig:xarm_v_dls}
    \end{subfigure}
    
    \caption{Characterization experiments on a simulated X-Arm7 reaching line keypoints listed in~\figref{fig:keypoints} with controller parameters~$\boldsymbol{K} = 10 \boldsymbol{I}, \Delta t = 0.02$~s ($50$~Hz controller). The region shaded in yellow corresponds to keypoint C, which is unreachable. The small, transient orientation errors are discussed in \secref{sc:disc}. Joint velocities are colored by algorithm, to avoid confusion (each color darkening with increasing joint index from~$q_1$ to~$q_7$). }
    \label{fig:xarmsim}
\end{figure*}

The shaded regions of the plots in~\figref{fig:xarmsim}~($20$~s each) show the different keypoints; \new{position} errors rise every time the goal pose changes (i.e., moves to a new keypoint). \new{Even though the same orientation is set for all keypoints, transient errors in orientation are expected, since J-PARSE does not decouple the control of translation and rotation~(\secref{sc:taskspacecoup}).}

First, J-PARSE is compared with the regular pseudoinverse, for the sake of completeness in \figref{fig:xarm_jparse_v_pinv}. As expected, the results match exactly away from singular regions, and pseudoinverse is unstable near singularities. In the plane keypoints case, the instabilities caused the simulation to crash. 

Comparisons of J-PARSE with its ablations in \figref{fig:abl1}, \figref{fig:abl2} show the role of each term. JacobianSafety~(\figref{fig:abl1}) \new{causes chattering}, as it allows singular motions to be commanded but simply inverts a constructed, and therefore inaccurate, Jacobian matrix. While there is no division by zero, the commanded task-space velocity in singular directions is still divided by a small number~$\gamma \sigma_{\mathrm{max}}$. JacobianProjection~(\figref{fig:abl2}) also leads to unstable behavior, as it has no mechanism to transition between singular and non-singular regions \ins{(therefore chattering when the ICN reaches~$\gamma = 0.1$) }, and as~$\boldsymbol{J}_p^+$ alone \ins{does not form a} projection matrix to \ins{correctly} eliminate the singular commands. JacobianSafetyProjection~(\figref{fig:abl2}) successfully eliminates the singular directions, and therefore remains stable, but having eliminated movements in those directions, it does not succeed in reaching the desired poses. 

J-PARSE compared with various damping values for DLS (\figref{fig:xarm_v_dls}) shows lower steady-state errors at non-singular poses compared to DLS with high damping, and stability in singular regions compared to DLS with low damping. This is an expected behavior of DLS with constant damping. As discussed in \secref{sc:2r}, adaptive damping is capable of navigating this tradeoff, but is difficult to tune. 

\begin{figure}[htbp]
    \centering
    \begin{subfigure}[b]{0.4\textwidth}
        \centering
        \includegraphics[width=\textwidth]{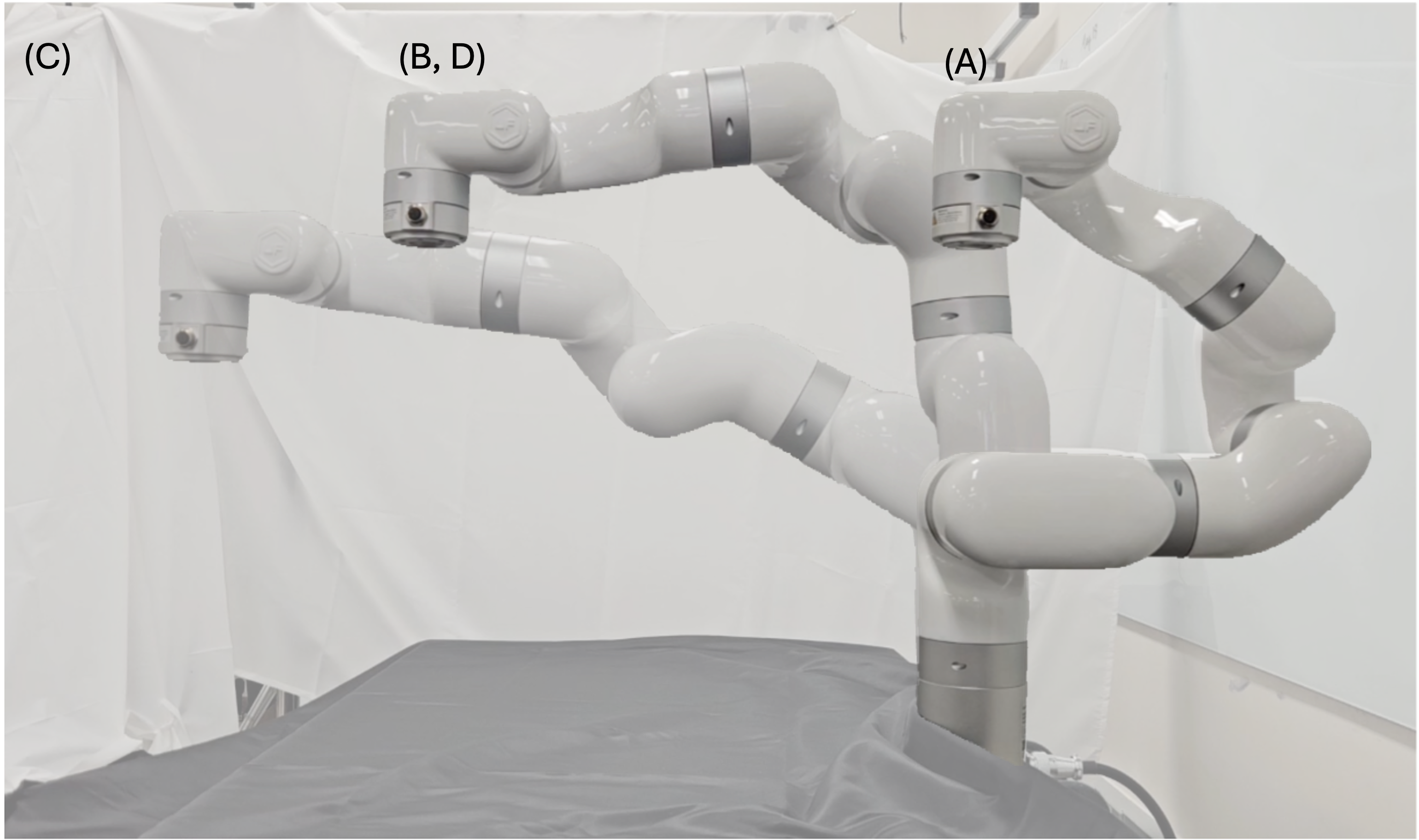}
        \caption{Physical X-Arm7 reaching towards line keypoints.}
        \label{fig:xarm_real_world_line_keypoints_photo}
    \end{subfigure}
    \begin{subfigure}[b]{0.45\textwidth}
        \centering
        \def\svgwidth{\textwidth} 
        {\sffamily
        \tiny
        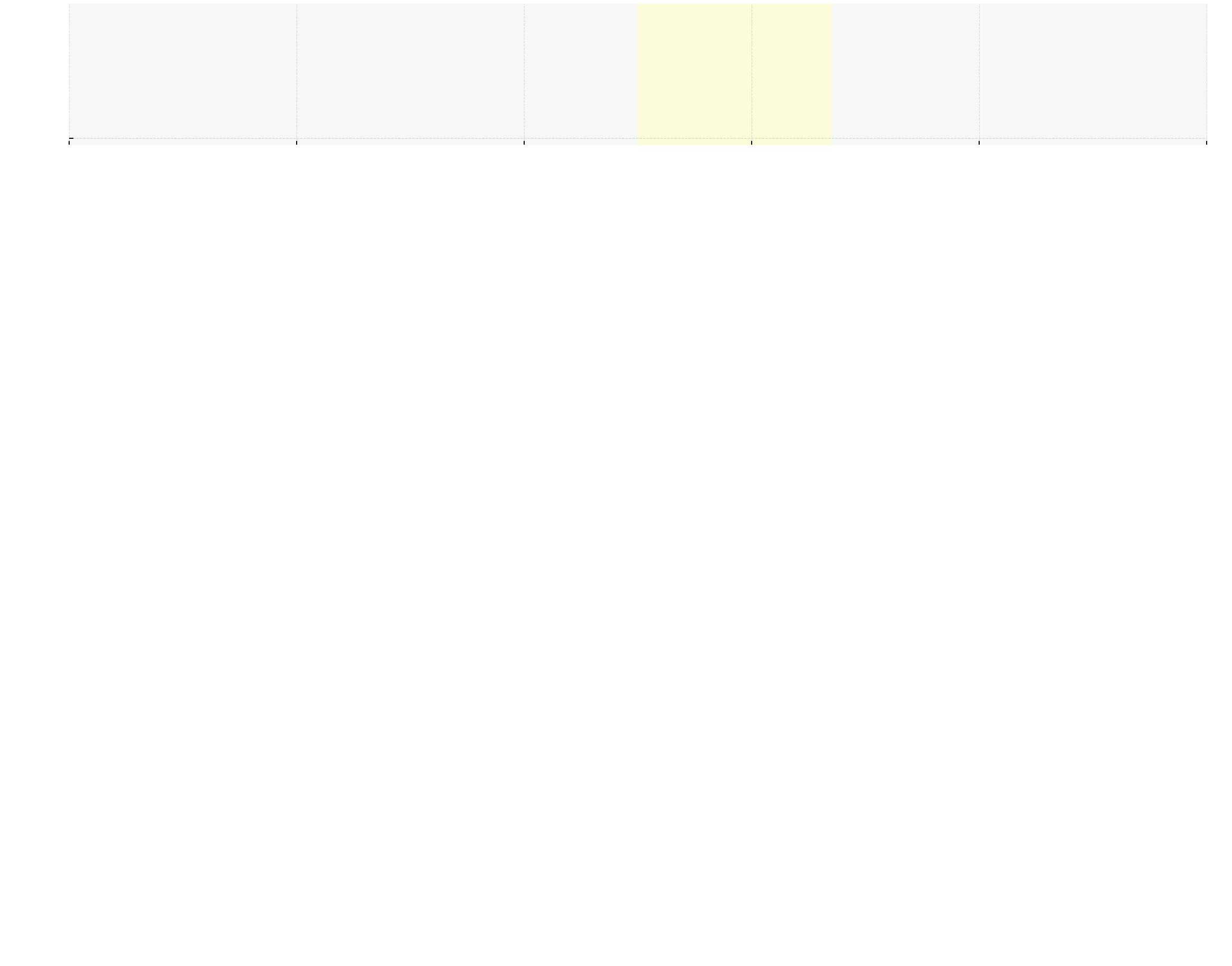
        \normalsize
        }
        \caption{Real world results with J-PARSE on Line Keypoints.}    
        \label{fig:real_xarm_line_ellipse_keypoints}
    \end{subfigure}
    \caption{Characterization experiments on a physical X-Arm7 reaching line keypoints listed in~\figref{fig:keypoints} with controller parameters~$\boldsymbol{K} = \boldsymbol{I}, \Delta t = 0.02$~s ($50$~Hz controller). The region shaded in yellow corresponds to keypoint C, which is unreachable. Due to the lower gains, the transient orientation errors are an order of magnitude lower compared to the simulation.}
    \label{fig:xarmreal}
    \vspace{-1em}
\end{figure}

\figref{fig:xarmreal} shows the same keypoints reached by the physical robot. The behavior was stable, as expected from simulations; videos are available as supplementary material. \ins{Sampling at~$10$~Hz, the peak joint speed reported by the driver over the cycle was~$0.70~\mathrm{rad/s}$~($40\degree$/s, well below the maximum recommended value~\cite{xarm_manual} of~$180\degree$/s) at joint~$4$. Peak acceleration and jerk values calculated from driver-reported velocities were~$3.40~\mathrm{rad/s^2}$~($194.90\degree~\mathrm{/s^2}$) and~$33.53~\mathrm{rad/s^3}$~($1922\degree~\mathrm{/s^3}$) respectively, both of which are below the maximum values recommended by the manufacturer~($1145\degree~\mathrm{/s^2}$ and~$28647 \degree~\mathrm{/s^3}$).}
\subsection{Internal singularities of spatial manipulators}
\label{sc:intsing}
The~$7$-\gls{DoF} Kinova Gen3 and~$6$-\gls{DoF} PUMA560 manipulators were used for tracking trajectories through singular configurations internal to the workspace. As extensive comparisons in the previous sections have established, DLS is able to achieve comparable movement to J-PARSE if appropriately tuned, is sluggish and has steady-state errors at high damping, and risks instability at low damping. The advantage of J-PARSE lies in safety, reliability and \new{ease of} tuning. Therefore, plots presented in this section demonstrate the use of J-PARSE in detail, without repeating the comparisons against other algorithms.

\begin{figure}[ht]
\centering
\def\svgwidth{0.53\textwidth} 
        {\sffamily
        \tiny
        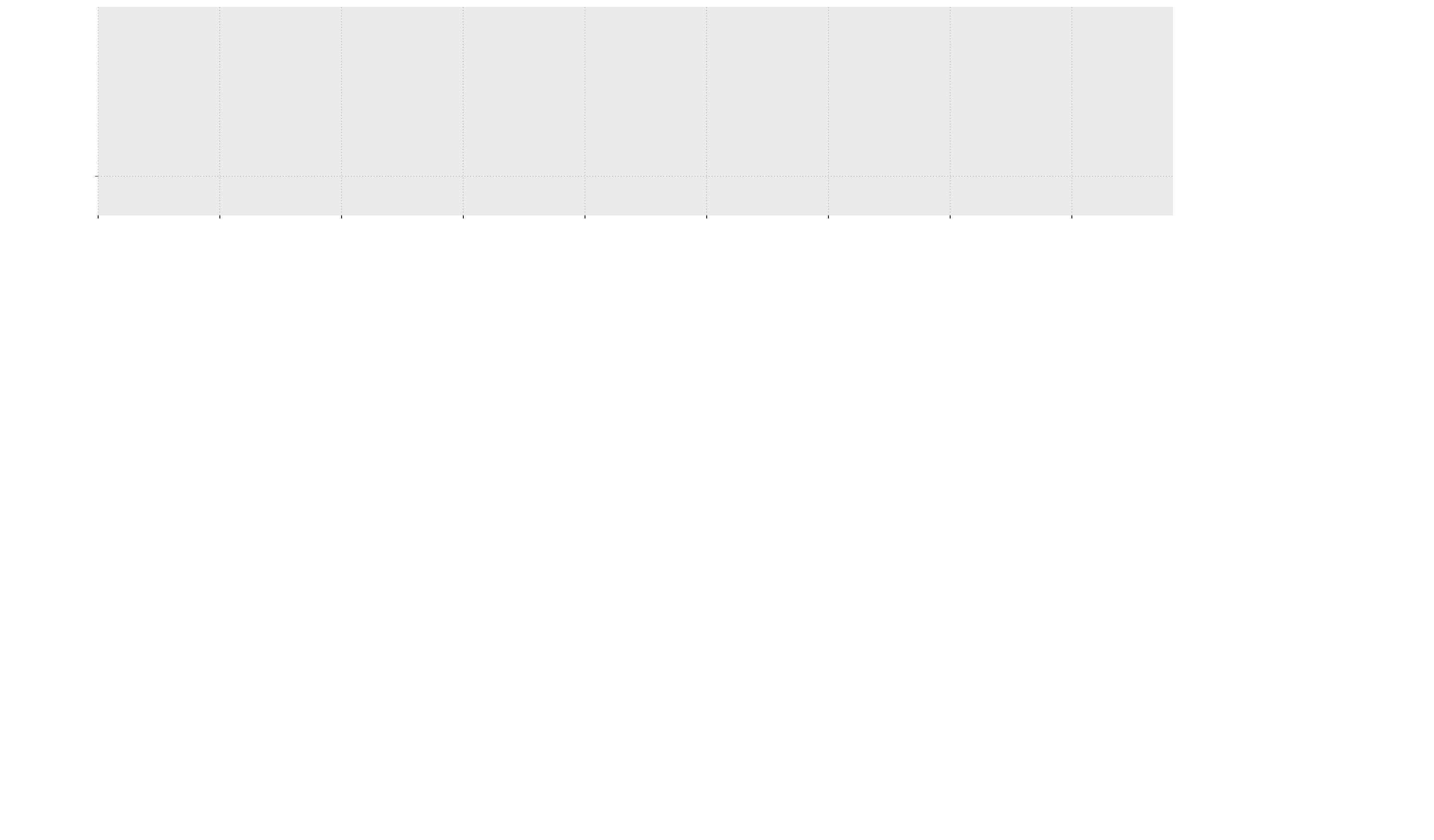
        }
        \vspace{-0.5em}
        \caption{\ins{Kinova Gen3 passing through and close to internal singularities with a human in the loop. The ICN drops to zero at both internal and boundary singularities. As expected with J-PARSE, joint velocities are well within reasonable limits throughout.~$k_x = k_y = 2, k_z = 1$; controller frequency~$30$~Hz.}}
        \label{fig:gen3traj}
\end{figure}

\begin{figure}[t]
     \centering
         \def\svgwidth{0.45\textwidth} 
        {\sffamily
        \tiny
        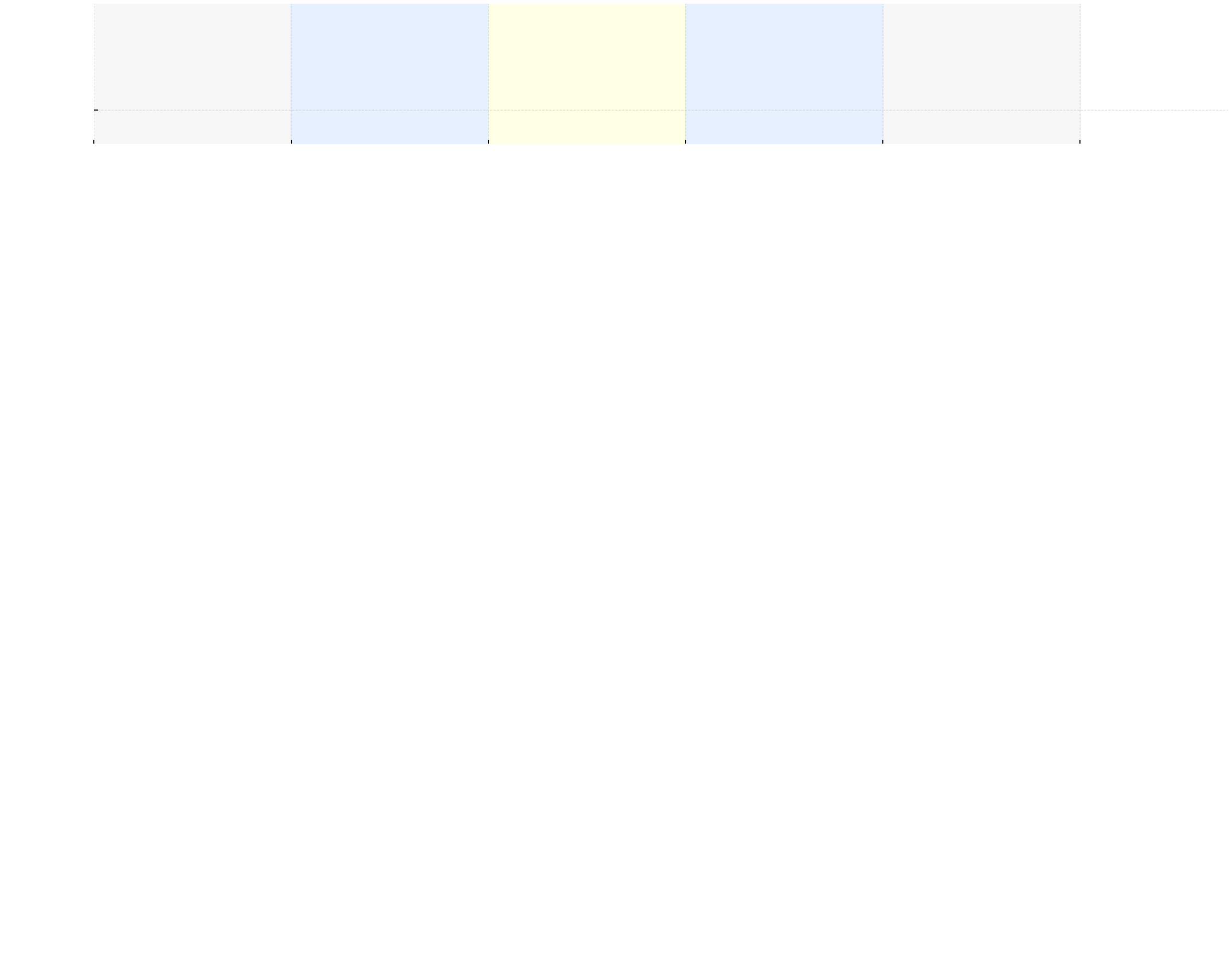
        }
        \caption{PUMA560 following an open-loop sinusoidal motion with two values of~$\gamma$, passing through gimbal lock. The plots show (from top to bottom) reference position coordinates (m), position error (m), orientation error (rad), ICN, and~$q_5$ (rad). At every dashed line marked on the plot, the manipulator is at the singular configuration shown at the top, with~$q_5$ crossing zero. As expected, behavior diverges at ICN of~$0.05$. Joint velocities are shaded darker with increasing joint index.}
        \label{fig:pumatraj}
        \vspace{-1em}
\end{figure}

As the robot must slow down in singular directions, a human teleoperator would \ins{adapt the desired twist command~$\boldsymbol{\mathcal{V}}$ in a closed loop, responding to the movement of the robot. For the Gen3 manipulator, since a physical robot was available, the end-effector was controlled in real time \new{using the 3D-Connection SpaceMouse\textregistered{}, which allows for the prescription of a~$6$-axis command twist~$\boldsymbol{\mathcal{V}}$ for the end-effector.} For the PUMA, the manipulator is simulated in Gazebo and a goal pose is published by a ROS node. A low speed is maintained throughout so that the goal pose (since it is published in an open-loop fashion, unlike a real application) does not move too far ahead of the end-effector as the end-effector slows down near the singularity.}  \\
\textbf{1) } \ins{The Gen3 was controlled at~$30$~Hz \new{and passed through singular configurations at various points in time. \figref{fig:gen3traj}) shows it reaching internal singularity when~$q_2$ crosses zero (first joint aligned with third), becoming near-singular again when~$q_2$ is zero and~$q_6$ is nearly zero (fifth joint nearly aligned with seventh), and reaching the boundary singularity when~$q_4 = 0$ and~$q_6$ is nearly zero).}. The task-space gains were set to~$k_{x} = 2, k_y = 2, k_z = 1, \new{k_{\omega x} = k_{\omega y} = k_{\omega z}} = 0.5$ ($k\Delta t = \frac{2}{30} \approx \frac{2}{31}$, barely exceeding the conservative bound\footnote{\new{The relaxed condition $k\Delta t \leq 2$ in \eqref{eq:conservativebound} holds only for isotropic gains~${\boldsymbol{K} = k\boldsymbol{I}}$; the Gen3 uses non-isotropic gains, so the applicable sufficient condition is the general bound~\eqref{eqn:discretecond},~$k\Delta t \leq \frac{2}{m(m-1)+1} = \frac{2}{31}$. The chosen $k\Delta t = \frac{2}{30}$ marginally exceeds this sufficient (not necessary) condition, and stable behavior is nonetheless observed.}} derived in \secref{app:suffcond}), with the singularity threshold~$\gamma = 0.1$}. \new{The manipulator encounters both internal and boundary singularities and motion remains stable throughout.}\\
\textbf{2)} For the PUMA, the target pose was slowly moved along a straight line in the~$y$ direction, such that that the manipulator passed through wrist-lock or gimbal lock (alignment of fourth and sixth joint axes) twice during each cycle. The path was defined by~${x = 0.432}$~m,~${z = 1.105}$~m, and~$y$ varying sinusoidally between~$\pm 0.3$~m, with constant orientation. Gains were set to~$k = 50$ for all \gls{DoF}, controlling at~$50$~Hz. Results are shown in \figref{fig:pumatraj}. \new{As expected, the manipulator slows down close to the singularity, and takes time to return to a higher-manipulability configuration. Due to the open-loop nature of the setup, this allows errors to rise for a period immediately following the crossing of singularity. Overall, errors are small in both position and orientation, and are smaller for the smaller value of~$\gamma$, as it causes deceleration during a smaller part of the trajectory and thus causes less delay in following the moving goal. }




\subsection{Practical examples of online control} 
\label{sc:demos}
This section reports demonstrations of J-PARSE for various applications on real manipulators in the full~$6$-\gls{DoF} task space, showing adaptive tracking of position and orientation with a human-in-the-loop.

\begin{figure*}[ht]
\centering
\def\svgwidth{0.7\textwidth} 
{\sffamily
\tiny
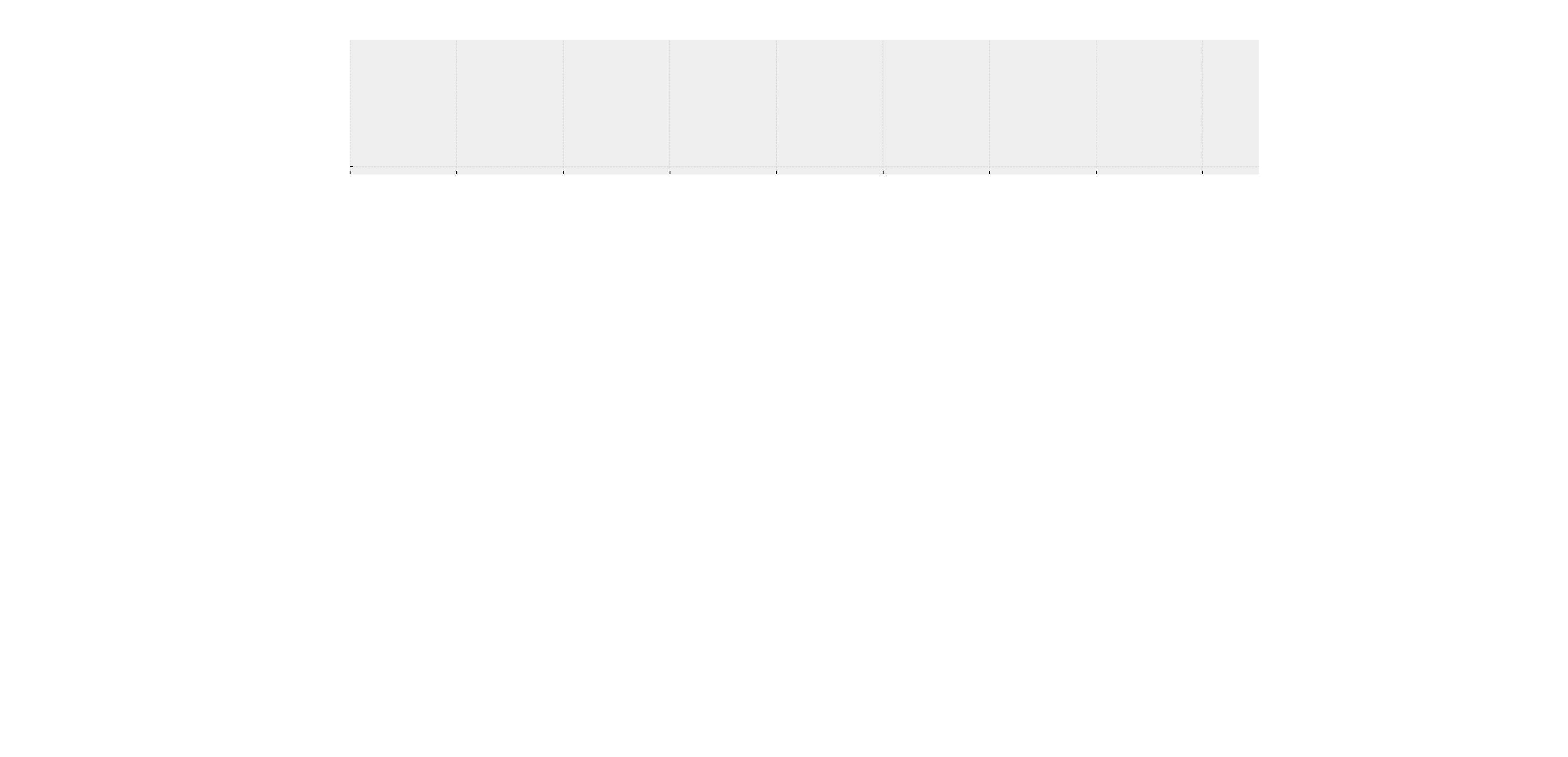
}
\caption{\ins{J-PARSE compared with the default X-Arm7 controller in a pick-and-place task for an object near the boundary of the workspace. End-effector twists are commanded via SpaceMouse (controller frequency~$30$~Hz). Before reaching the object, the default controller causes joint speeds to shoot up, reports an error and severs the connection. \new{With J-PARSE, the task is successfully completed.}}}
\label{fig:xarm_teleop}
\vspace{-1em}
\end{figure*}

1) \textbf{Teleoperation}: Teleoperation of the X-Arm was performed using \new{a }SpaceMouse\textregistered{}. The human demonstrator was given the task of picking a green cup near the boundary of the manipulator's workspace and placing it near the base. The performance of the default \ins{Cartesian }controller provided by the manufacturer is compared with J-PARSE in~\figref{fig:xarm_teleop}. User inputs were scaled along each~\gls{DoF} to a range of~$[0, 0.08]$ for position and~$[0, 0.3]$ for orientation, tuned by operator preference. Using J-PARSE, the arm is controlled to retrieve the object successfully. Anecdotally, demonstrators found teleoperation easy to use. A formal study on its intuitiveness is left as future work. \label{sc:teleop_results}

\begin{figure*}[ht]
\centering
\def\svgwidth{0.7\textwidth} 
{\sffamily
\tiny
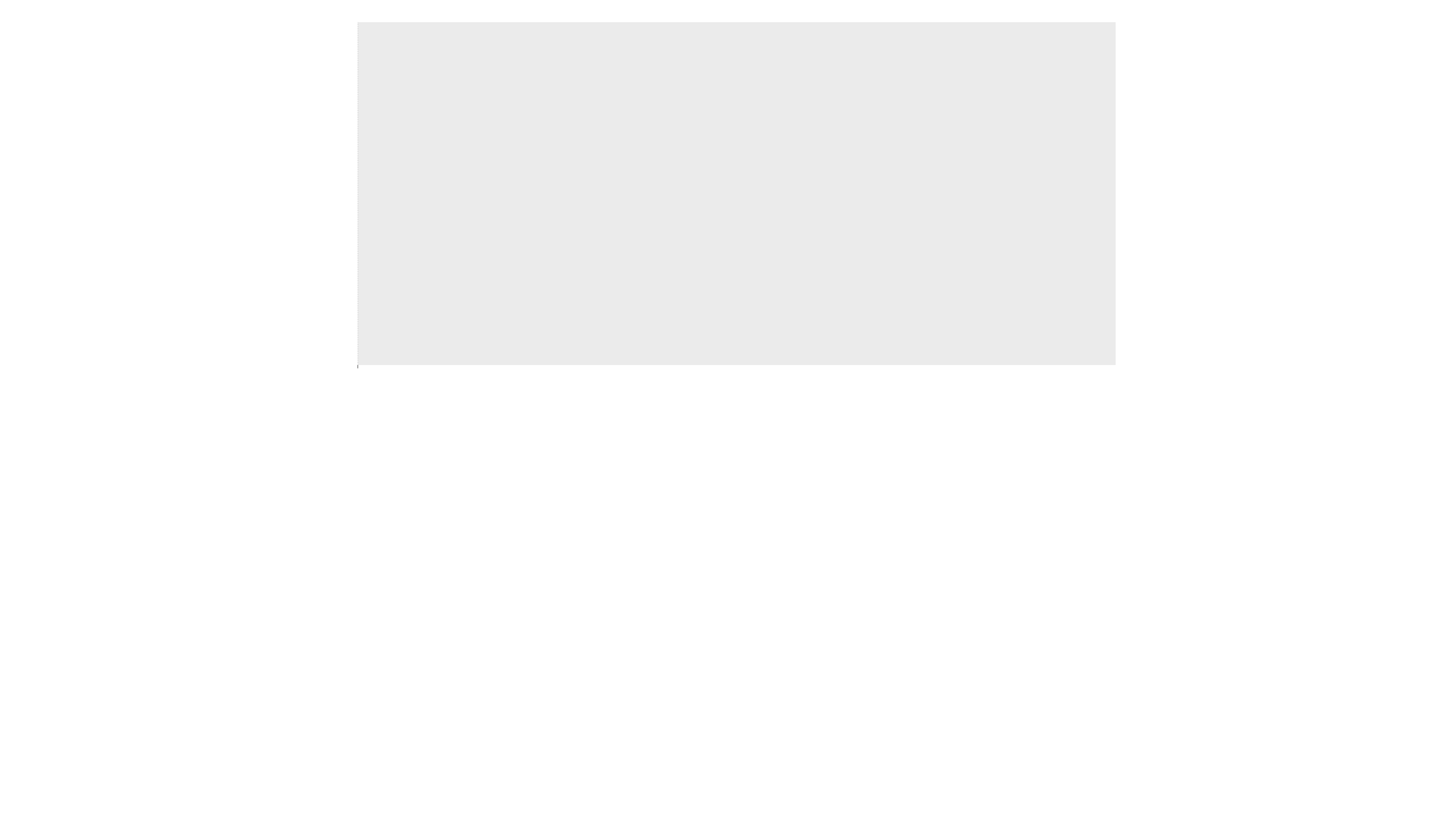
}
\caption{\ins{Visual servoing with the Gen3 manipulator, causing it to enter a singular configuration to reach towards a target pose outside the workspace, and then retract to a non-singular region, with joint velocities remaining within reasonable bounds throughout;~$k_x = k_y = 2, k_z = 1$; controller frequency~$30$~Hz.}}
\label{fig:visual_servo}
\vspace{-1em}
\end{figure*}

2) \textbf{Visual servoing}: As an example of continuous following of a goal pose, the Intel RealSense D415 camera integrated at the end-effector of the Kinova Gen3 manipulator was used to track a pose facing a hand-held fiducial (AprilTag~\cite{olson2011tags}) at an offset of~$0.50$~m. \figref{fig:visual_servo} shows the manipulator reaching towards and retracting from a distant goal, with all motion being performed in one continuous interaction.

3) \textbf{Learning from Demonstration}: For learning from demonstration, the \gls{DDPM} framework~\new{\cite{hoDenoisingDiffusionProbabilistic2020a,chi2024diffusion}} is used to emulate the human demonstrator, who retrieves the green cup from the boundary of the workspace using the X-Arm with velocity control, as in~\secref{sc:teleop_results}-1. This serves as a proof of concept that J-PARSE can be quickly adapted to learning frameworks, making it possible for these trained models to operate near manipulator singularities. Diffusion Policy~\cite{chi2024diffusion} is a state-of-the-art imitation learning algorithm that is built on DDPM. The formulation is as follows: the observation of the model is an RGB image of the robot $I_{\mathrm{RGB}}$, robot pose $\boldsymbol{p}$ (in which the rotation is often converted to a $6$-D representation), and the output is the task-space pose command vector $\boldsymbol{a}$, which can be converted to commanded twist $\boldsymbol{\mathcal{V}}$ for velocity control. The probabilistic model solved by \gls{DDPM} is the conditional probability \new{$P(\boldsymbol{a}_k| I_{\mathrm{RGB},k})$} (for current time \new{index} $k$). For denoising steps $d \in [1,D]$, the iterative denoising is:
\begin{equation}
\boldsymbol{a}_{k}^{d-1} = \alpha_1(\boldsymbol{a}_k^{d} - \alpha_2 \epsilon_{\theta}(I_{\mathrm{RGB},k},\boldsymbol{a}_k^{d},d) + \mathcal{N}(0,\sigma^2 I)),
\label{eqn:diffusion_denoising}
\end{equation}
where $\mathcal{N}(0,\sigma^2I)$ is Gaussian noise added at each iteration with variance $\sigma$, $\epsilon_{\theta}$ is the noise prediction network with parameters $\theta$, and $\alpha_i$ are noise schedule gains. The model training \gls{MSE} loss is 
\begin{equation}
    \mathcal{L} = \mathrm{MSE}({\epsilon}^d,\epsilon_{\theta}(I_{\mathrm{RGB},k},\boldsymbol{a}_k^{d},k))
\end{equation}
and the $\epsilon_{\theta}(\cdot)$ network architecture is a 1D temporal CNN that features $I_{\mathrm{RGB}}$ observations using \gls{FiLM} \cite{perezFiLMVisualReasoning2018}, as in \cite{chiDiffusionPolicyVisuomotor2024}.

\ins{A total of~$74$ successful demonstrations from the human was used to train} the imitation learning model. For each demonstration, a human teleoperates the X-Arm to pick up a green cup \ins{near the boundary of its reachable workspace} (entering singularity), and moves it to deposit the cup in a specified region (exiting singularity). The position of the cup at the pick-up and drop point is slightly varied (up to~$5$~cm away from a center point) to encourage generalization in the policy. Future work will include picking up different objects from completely different singular regions.

The robot pose (comprising a~$3$-D position,~$6$-D rotation, and~$1$-D gripper position), the $10$-D pose action, and an RGB image from the Intel Realsense D435i are collected \ins{for each demonstration}.
Color jittering is performed on the images to make the policy robust to lighting perturbations, and the parameters are kept the same as the original RGB-based Diffusion Policy. Training is conducted on a Nvidia RTX 5090 GPU for approximately $20$ hours to $2000$ epochs. The policy operates at $10$~Hz on a Nvidia RTX 5090 GPU. During inference, the Diffusion Policy predicts pose actions, which are converted to desired end-effector velocities, and sent to J-PARSE to output joint velocities. 

\begin{figure}[ht]
    \centering
\def\svgwidth{0.48\textwidth} 
{\sffamily
\fontsize{3.5}{5}\selectfont
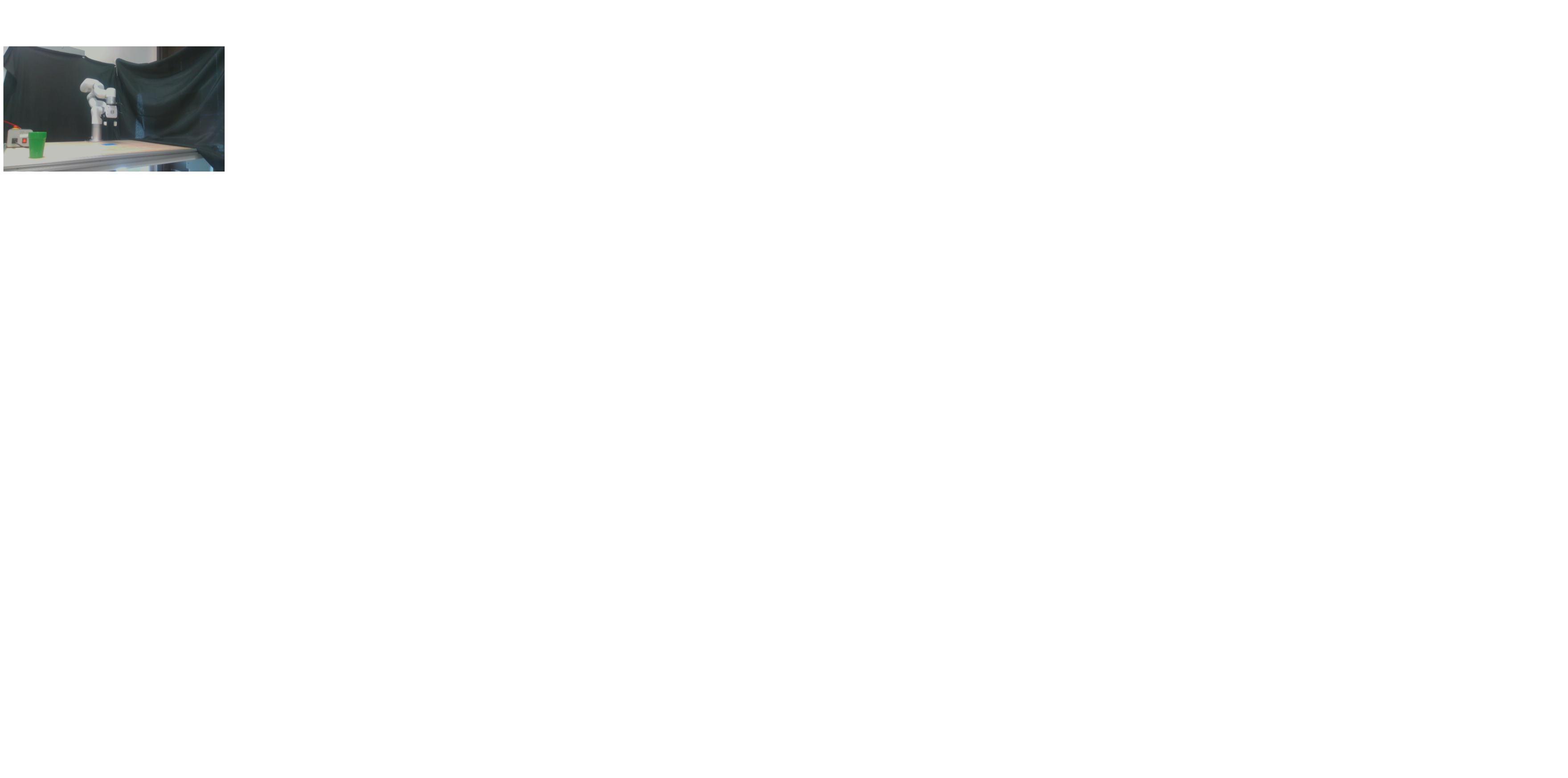
}
    \caption{Imitation Learning. A human provides~$74$ demonstrations of commanded task-space twist ${\boldsymbol{\mathcal{V}}}$; then the robot learns to condition the robot pose action, which is converted to a twist action, on the third-person camera view from that data. Above, a successful single-action demonstration is shown as proof of concept for the use of J-PARSE for imitation learning.}
    \label{fig:cartesian_lfd}
    \vspace{-1em}
\end{figure}

\figref{fig:cartesian_lfd} shows the robot policy successfully retrieving a green object from the boundary of the workspace and returning it to a specified goal region, which requires \emph{entering and exiting a singular region}. \new{The policy was successful in~$9$ out of~$10$ trials.} \new{This proof of concept indicates} that robust and safe policies can be learned where the robot may be required to use its full workspace to perform a task near singularity, \new{making J-PARSE a viable option for learning algorithms}.

\section{Discussion}
\label{sc:disc}
In \new{IK} control, the design of methods for handling singularity is, by necessity, a process of choosing the least of several evils (Table~\ref{tab:qualcomp}), and providing a tunable behavior between two tradeoffs. For \gls{DLS}, the tradeoff is between stability near singularities and accuracy elsewhere; for other methods which explicitly avoid singular regions, the tradeoff is between stability and reach. In every method, the commanded end-effector velocity is modified -- in either direction, or magnitude, or both -- when in the vicinity of a singularity. J-PARSE makes this modification in a manner that naturally conforms to the kinematics; that is, the direction and speed are modified \emph{to the extent that the request is unreasonable otherwise}. \new{It does so with a single tuning parameter. The cost of this ease of tuning is a discontinuity introduced in joint accelerations at the~$\kappa_{\mathrm{cn}} = \frac{1}{\gamma}$ boundary, which does not compromise stability either in theory or practice.} 

\ins{In the modified form of J-PARSE proposed in \secref{sc:mods}}, the parameters~$\gamma$ and~$a$ together control the severity and nature of the modification of the commanded velocity. Some considerations for their selection have been presented. In future work, it would be of interest to further develop principled methods for the selection and online tuning of these parameters based on kinematic architecture, joint position, speed, and acceleration limits, approach to or retreat from singularities, and desired constraints upon the motion in task space. 

\new{An important limitation of the algorithm is its entirely local nature. }It is assumed \new{in this work }that there \new{exists} a \new{known} sequence of feasible \new{configurations} connecting the current and goal pose. J-PARSE \new{is intended to be used only to compute} desired joint speeds from \new{the} desired end-effector velocity\new{, with consideration of the available mobility. It makes singularity avoidance unnecessary, but} does not \new{account} for the additional constraints of \new{avoiding} self-collision, external collisions and joint limits. It also does not keep position and orientation decoupled (\secref{sc:mods}), and if large changes in orientation are commanded at once, it may result in rapid motions, due to the non-Euclidean nature of~$\sothree$. 

\new{Moreover, even though the stability conditions are agnostic to the manipulator architecture, the nominal configuration used to design the nullspace term needs to be selected by the designer based on practical considerations. In real-time human-in-the-loop operation, it is not to be assumed that the nullspace term (in its current form, with a constant nominal pose) is a sufficiently intuitive approach for resolving redundancy. Nor does the current work establish that human operators prefer the use of J-PARSE over EDLS, or that human designers prefer the tuning of J-PARSE over that of ADLS. The relevant user studies and design improvements are left for future work.}

\new{In the autonomy or human-in-the-loop stack,} J-PARSE \new{can be integrated as a layer} between task\new{-space} velocity commands and joint velocity commands. \new{Given} the kinematic \new{architecture} of the robot, J-PARSE \new{provides} an easy-to-implement, reliable method to ensure safe motion in and out of singular configurations. This becomes useful in more expressive robot policies, teleoperation, and control of robots around human beings, for example, in medical or assistive settings.

\section{Conclusion}
\label{sc:conc}
This work presents J-PARSE, an \new{IK} control algorithm for maneuvering manipulators \new{reliably throughout their reachable workspace, \textbf{without avoiding singularities}, without requiring \textit{a priori} analysis of singularities or dynamics, and \textbf{without context-dependent tuning of multiple parameters}. It guarantees local stability agnostic to the presence or proximity of singularities, using a single kinematically-meaningful parameter. The insight that enables such behavior is that modern applications of online robotic control present a strong motivation for prioritizing a high degree of \emph{capability} throughout the workspace, over accurate tracking of fast trajectories. Future work includes evaluating the ease of tuning and teleoperation through user studies, using J-PARSE in combination with collision-avoidance and path planning algorithms, and further developing the proposed variations of the algorithm.}
%
%
\appendix
\section{Sufficient condition for stable selection of gains}
\label{app:suffcond}
From the discrete-time analysis in \secref{sc:discproof}, the condition for stability is the positive semi-definiteness of~$\boldsymbol{\Theta}$, defined by: 
\begin{align}
    \Theta_{ij} &= Q_{ij} - \frac{\Delta t}{2} Q_i Q_j \sum_{l=1}^m{K_l U_{lj} U_{li}} \\
    &= \begin{cases}
                        - \frac{\Delta t}{2} Q_i Q_j \sum_{l=1}^m{K_l U_{lj} U_{li}} & \text{if } i \ne j  \\
                        1 - \frac{\Delta t}{2} \sum_{l=1}^m{K_l U_{li}^2} & \text{if } i = j \textrm{ and} \\ & \sigma_i \geq \gamma \sigma_{\mathrm{max}} \\
                        \frac{\sigma_i^2}{\gamma^2 \sigma_{\mathrm{max}}^2} \left(1 - \frac{\Delta t}{2} \frac{ \sigma_i^2}{\gamma^2 \sigma_{\mathrm{max}}^2} \sum_{l=1}^m{K_l U_{li}^2 }\right)  & \text{if } i = j \textrm{ and} \\ & \sigma_i < \gamma \sigma_{\mathrm{max}} 
    \end{cases},
    \label{eqn:messy_cases}
\end{align}
\ins{where~$\boldsymbol{Q}$ is defined as in~\eqref{eqn:Qdef}.} As~$\Theta$ is a symmetric matrix, a sufficient condition for~$\Theta$ to be positive semi-definite is that it is diagonally dominant and all its diagonal entries are non-negative~\cite{johnston2021advanced}. A conservative approach is taken below, to identify a subset of parameters satisfying this condition. 

Let~$k$ be an upper-bound on the entries of the gain matrix~$\boldsymbol{K}$. Then,~${\sum_{l=1}^m K_l U_{li}^2 \leq \sum_{l=1}^m k U_{li}^2 = k}$. It follows that the diagonal elements of~$\boldsymbol{\Theta}$ are non-negative if~${k \Delta t \leq 2}$.

The matrix~$\boldsymbol{\Theta}$ is said to be diagonally dominant iff the sum of absolute values of its off-diagonal entries in each row does not exceed the absolute value of the diagonal entry in that row. When diagonal entries are non-negative, the following is an equivalent statement:
\begin{align}
    \sum_{j \neq i, j = 1}^m|\Theta_{ij}| \leq \Theta_i, \forall i = 1, \dots, m.
\end{align}

For all~$i$ such that~$\sigma_i < \gamma \sigma_{\mathrm{max}}$, the condition is:
\begin{align}
    &\sum_{j \neq i, j = 1}^m \left| \frac{\Delta t}{2} \frac{\sigma_i^2}{\gamma^2 \sigma_{\mathrm{max}}^2} Q_j \sum_{l = 1}^m K_l U_{lj} U_{li} \right| \leq \\
    &\frac{\sigma_i^2}{\gamma^2 \sigma_{\mathrm{max}}^2} \left( 1 - \frac{\Delta t}{2} \frac{\sigma_i^2}{\gamma^2 \sigma_{\mathrm{max}}^2} \sum_{l=1}^m K_l U_{li}^2 \right),
\end{align}
where~$Q_j$ takes values~$1$ or~$\frac{\sigma_i^2}{\gamma^2 \sigma_{\mathrm{max}}^2}$, depending on whether~$\sigma_j$ is greater than~$\gamma \sigma_{\mathrm{max}}$, or not. Thus, the greatest possible value of the left-hand side would be achieved if~$\sigma_j > \gamma \sigma_{\mathrm{max}} \forall j \neq i$. Therefore, it is sufficient that:
\begin{align}
    \label{eqn:maininequality}
    \sum_{j \neq i, j = 1}^m \left| \frac{\Delta t}{2} \sum_{l = 1}^m K_l U_{lj} U_{li} \right| \leq 1 - \frac{\Delta t}{2} \frac{\sigma_i^2}{\gamma^2 \sigma_{\mathrm{max}}^2} \sum_{l=1}^m K_l U_{li}^2,
\end{align}
as~${ \frac{\sigma_i^2}{\gamma^2 \sigma_{\mathrm{max}}^2} \geq 0}$. A upper bound on the left-hand side is identified by recognizing that~${U_{li}, U_{lj} \leq 1}$:
\begin{align}
    \sum_{j \neq i, j = 1}^m \left| \frac{\Delta t}{2} \sum_{l = 1}^m K_l U_{lj} U_{li} \right| \leq (m-1) \frac{ \Delta t}{2}(m k),
\end{align}
while a lower bound on the right-hand side is found by maximizing the negative term, 
\begin{align}
     1 - \frac{\Delta t}{2} \frac{\sigma_i^2}{\gamma^2 \sigma_{\mathrm{max}}^2} \sum_{l=1}^m K_l U_{li}^2 \geq 1 - \frac{\Delta t}{2} k.
\end{align}

A conservative sufficient condition for diagonal dominance is that the upper bound on the left-hand side of~\eqref{eqn:maininequality} does not exceed the lower bound on the right-hand side:
\begin{align} 
    (m-1)~m~k~\frac{ \Delta t}{2} &\leq 1 - \frac{\Delta t}{2} k \\
    \Rightarrow \quad k \Delta t &\leq \frac{2}{m(m-1) +1}.     \label{eqn:discretecond}
\end{align}

Since~$m \geq 1$,~\eqref{eqn:discretecond} is a stricter condition than~${k \Delta t \leq 2}$. Following a similar procedure for all~$i$ such that~$\sigma_i \geq \gamma \sigma_{\mathrm{max}}$, it is easily verified that~\eqref{eqn:discretecond} also ensures the diagonal dominance condition is satisfied for those rows. Therefore,~\eqref{eqn:discretecond} ensures that~$\boldsymbol{\Theta}$ is positive semi-definite, and may be used as a guideline for selection of~$\boldsymbol{K}$ based on~$m$ and~$\Delta t$. For example, for~${m=6}$ and~${\Delta t = 0.01}$, the proportional gains should be upper-bounded by approximately~$6.45$. Future work will attempt to identify less strict conditions that may also guarantee stability.

\bibliographystyle{./IEEEtran}
\bibliography{bib/jparse_kennedy, bib/jparse_guptasarma}

\end{document}